\documentclass[12pt]{article}
\usepackage{multirow}
\usepackage{graphicx}
\usepackage{graphics}
\usepackage{epsfig}
\usepackage[below]{placeins}
\usepackage{subfigure}
\usepackage{float}
\usepackage{rotating}
\usepackage{amsmath}
\usepackage{longtable}
\usepackage{booktabs}
\usepackage{subfigure}

\begin{document}

\begin{center}

{\bf \Large A New Path to Construct Parametric Orientation Field: Sparse FOMFE Model and Compressed Sparse FOMFE Model}

\bigskip

{\bf Jinwei Xu, Jiankun Hu, Xiuping Jia}\\

School of Engineering and Information Technology\\

The University of New South Wales\\

Canberra, ACT 2600, Australia\\

E-mails: jinwei.xu@student.adfa.edu.au; j.hu@adfa.edu.au; x.jia@adfa.edu.au

\end{center}

\begin{center}
{\bf Abstract}
\end{center}

Orientation field, representing the fingerprint ridge structure direction, plays a crucial role in fingerprint-related image processing tasks. Orientation field is able to be constructed by either non-parametric or parametric methods. In this paper, the advantages and disadvantages regarding to the existing non-parametric and parametric approaches are briefly summarized. With the further investigation for constructing the orientation field by parametric technique, two new models - sparse FOMFE model and compressed sparse FOMFE model are introduced, based on the rapidly developing signal sparse representation and compressed sensing theories. The experiments on high-quality fingerprint image dataset (plain and rolled print) and poor-quality fingerprint image dataset (latent print) demonstrate their feasibilities to construct the orientation field in a sparse or even compressed sparse mode. The comparisons among the state-of-art orientation field modeling approaches show that the proposed two models have the potential availability in big data-oriented fingerprint indexing tasks.

\bigskip

{\it Keywords:} fingerprint image, orientation filed, sparse representation, compressed sensing, FOMFE model, fingerprint indexing

\section{Introduction}

Orientation field (OF) is an important feature to represent the direction of ridge structures in a fingerprint image. To be specific, OF is applied to tune the Gabor filter directional parameters to enhance the ridge structures. Accordingly, the detailed feature such as minutiae is most likely to be accurately extracted based on the enhanced ridge segments. However, numerous fingerprint images in the real world are more or less contaminated by structural noise or incompletely collected (like latent print), therefore, the image quality are not satisfactory for the subsequent processing. In order to improve the fingerprint image quality for further analysis, specifically to connect the discontinuous ridge pattern and to recover the missing ridge segment, the OF-based contextual filtering is essential and performed as a pre-processing step prior to the subsequent fingerprint analysis tasks (like minutiae extraction, indexing, and matching).

OF modeling methods can be categorized as two following groups: (i) the non-parametric models; and (ii) the parametric models. The non-parametric OF modeling techniques have been intensively studied during the past decades, and consequently a large number of literatures can be found. Among such plenty of works, the representative ones are selected as state-of-art approaches and then further analyzed in this paper. A directional filtering scheme is proposed for ridge structure enhancement purpose \cite{Hong1998}. Therein, partial derivatives along $x-$ and $y-$ axis are calculated, and then applied to estimate the real ridge orientation. Such method can work well for the high-quality fingerprint images (e.g. rolled and plain prints), however, for the poor-quality ones (e.g. latent prints), the proposed method is not capable to yield the reliable orientation information. In order to strengthen the robustness to the structural noise, the orientation estimation is conducted by short time Fourier transform (STFT) supported by the fix-size sliding window \cite{Chikkerur2007}. The local Fourier analysis is able to capture the frequency response for the ridge pattern, therefore, the dominant frequency components in Fourier magnitude spectrum offers a powerful tool for ridge orientation estimation. The algorithm presented in \cite{Chikkerur2007} is working based on the following assumption: even the structural noise corruption could be severe, the ridge structure had to be the dominant one so that the Fourier spectrum could be still able to capture the dominant frequency component corresponding to the dominant ridge pattern. However, such assumption is nor likely to be guaranteed for the poor-quality fingerprint images. For further resisting to the structural noise, an OF template dictionary is established depending on the high-quality fingerprint image patches, and then utilized to correct the inaccurate ridge OF \cite{Feng2013}. Such technique incorporates the advanced data-driven learning strategy (dictionary learning) for constructing the OF template library, and then applies the learned OF templates to replace the incorrect ones according to the context-based similarity. However, the context-based similarity computation is not robust but vulnerable to the spurious OF information attained by coarse OF estimation based on the raw poor-quality latent fingerprint images. Therefore, the reliability of OF correction obtained by \cite{Feng2013} is still unsatisfactorily.

In contrast to the non-parametric approaches, the parametric models have the following advantages:

\begin{itemize}

  \item The modeling procedure only involves and depends on the limited ridge segments. In other word, for the poor-quality fingerprint images (like latent prints), it is unnecessary to select many ridge segments for OF modeling. Conversely, only a few salient ridge segments (the image segments with good-quality ridge structures) are sufficient and then be chosen for reliable OF modeling;

  \item The parametric model is capable to correct the initially estimated spurious OF information;

  \item The parametric model is performed as a guide to independently generate the orientation information for the unknown image segments. Particularly, for the low-clarity ridge segments, the parametric model is available to yield the OF in such segments, instead of directly estimating the unreliable orientation information according to these poor-quality ridge image segments.

\end{itemize}

According to the published works, the pioneering work can be traced back to \cite{Wang2007}. Therein, the fingerprint OF is the first time to be modeled based on a Fourier-like basis (namely FOMFE model). Based on FOMFE model, the initially estimated OF can be automatically corrected and visually fitted to the real ridge structure orientation. Further, the capability for FOMFE model to fill up the orientation information in missing image regions is developed \cite{Wang2011}. According to such foundation-stone works, diverse basis (transforms) are adopted to replace the original Fourier-like basis introduced in \cite{Wang2007} and \cite{Wang2011}. To be explicit, instead of adopting original Fourier-like basis, Legendre polynomials basis is utilized for OF modeling \cite{Ram2010}, and later discrete cosine transform (DCT basis) is also used \cite{Liu2014}. Besides, relying on the modeled OF, an invariant OF descriptor based on polar complex moment (PCM) is presented for rotation invariant  fingerprint indexing task \cite{Liu2012}. Although the currently developed parametric OF models have achieved the satisfactory performance in OF reconstruction, singular point positioning, fingerprint indexing and matching tasks, the yielded model coefficients are not sparse but dense. The dense nature of OF model coefficients will lead to the low searching efficiency and high storage consumption when dealing with the large-scale fingerprint indexing task in the context of the big-data. Such disadvantages are further analyzed as follows:

\begin{itemize}

  \item During fingerprint indexing task, the model coefficient is usually adopted as the feature vector to represent the individual fingerprint in feature space. Accordingly, the indexing task is conducted to find the most similar feature vectors (coefficient vectors) related to the given query fingerprint feature vector. Due to the dense nature of the feature vector, the similarity-based searching on one-to-one basis against the whole background database could be inefficient. That is, the similarity between candidate print feature vector and query print feature vector needs to be measured element-by-element (element means feature vector value, say if one feature vector has $121$ non-zero values (elements) then its $121$ non-zero values need to be involved into the feature vector similarity calculation);

  \item For storing the massive feature vectors contained in large-scale background database, a large volume storage space is essential, because all non-zero values in dense feature vectors need to be stored.

\end{itemize}

On the contrary, if the OF model coefficient (feature vector) could be as sparse as possible (most vector values are zeros), the one-to-one searching would be more efficient and the consumed storage space would be significantly reduced. The advantages for sparse OF model are summarized as follows:

\begin{itemize}

  \item The searching on one-to-one basis against a large-scale background database could be comparatively speedy. That is, it is unnecessary to measure the similarity over all feature vector elements, instead, only the non-zero values are considered. Consequently, when comparing with a query print feature vector, only the candidates whose non-zero element locations are the same as the query ones are selected for similarity measurement. Otherwise, the feature vectors whose non-zero locations are not the same as the query ones are directly filtered out.

  \item The storage space required for storing the massive feature vectors in the large-scale background database would be significantly reduced, since only the non-zero values in each feature vector need to be recorded.

\end{itemize}

In the consideration of the advantages of sparse OF model, in this paper, two new models - sparse FOMFE model (s-FOMFE) and compressed sparse FOMFE model (cs-FOMFE) are proposed respectively. The reminder of this paper is organized as follows: in Section \uppercase\expandafter{\romannumeral2}, the s-FOMFE and cs-FOMFE models are introduced; in Section \uppercase\expandafter{\romannumeral3}, the OF construction experiments on rolled fingerprint images and latent fingerprint images are conducted, and also the visual comparisons with the classical FOMFE model are demonstrated; in Section \uppercase\expandafter{\romannumeral4} the conclusion and future application regarding to the potential availability of the proposed s-FOMFE and cs-FOMFE models are given.

\section{Proposed Models}

Inspired by the rapid development and successful applications of signal sparse representation and compressed sensing theories in recent years, in this paper, two new OF models: sparse FOMFE model (s-FOMFE) and compressed sparse FOMFE model (cs-FOMFE) are introduced.

\subsection{Classical FOMFE Model (FOMFE)}

The classical FOMFE-based OF modeling is mathematically formulated as follows:

\begin{equation}
\hat \beta_{cos}^{FOMFE} = \mathop {\arg \min }\limits_{\beta_{cos}}  \left\| {b_{cos} - \Phi \beta_{cos} } \right\|_2^2
\label{classical_FOMFE_cos}
\end{equation}

\begin{equation}
\hat \beta_{sin}^{FOMFE} = \mathop {\arg \min }\limits_{\beta_{sin}}  \left\| {b_{sin} - \Phi \beta_{sin} } \right\|_2^2
\label{classical_FOMFE_sin}
\end{equation}
where $\Phi$ is a basis matrix and generated over every sampling point. $b_{cos}$ and $b_{sin}$ are single column observation vectors for $cos(2\Theta)$ and $sin(2\Theta)$, where $\Theta$ is a coarse orientation matrix over every sampling point. $\Theta$ can be obtained by using many local image patch-based orientation estimation techniques (here a conventional gradient-based method \cite{Maltoni2009} is employed). $\beta_{cos}$ and $\beta_{sin}$ are the FOMFE model coefficient vectors and needed to be solved based on Eqs. (\ref{classical_FOMFE_cos}) and (\ref{classical_FOMFE_sin}) respectively.

Such coefficient calculation problems can be converted to the following classical linear least square (LS) data-fitting problems:

\begin{equation}
\min \left\{ {\sum\limits_x {\sum\limits_y {\left\| {{b_{\cos }}\left( {x,y} \right) - \Phi \left( {x,y} \right) \cdot {\beta _{\cos }}} \right\|_2^2} } } \right\}
\label{LS_FOMFE_cos}
\end{equation}

\begin{equation}
\min \left\{ {\sum\limits_x {\sum\limits_y {\left\| {{b_{\sin }}\left( {x,y} \right) - \Phi \left( {x,y} \right) \cdot {\beta _{\sin }}} \right\|_2^2} } } \right\}
\label{LS_FOMFE_sin}
\end{equation}

The minimization of Eqs. (\ref{LS_FOMFE_cos}) and (\ref{LS_FOMFE_sin}) leads to the solutions for $\beta_{cos}$ and $\beta_{sin}$ as follows:

\begin{equation}
\hat \beta_{cos}^{FOMFE} = {\left( {{\Phi^T}\Phi} \right)^{-1}}{\Phi^T}{b_{cos}}
\end{equation}

\begin{equation}
\hat \beta_{sin}^{FOMFE} = {\left( {{\Phi^T}\Phi} \right)^{-1}}{\Phi^T}{b_{sin}}
\end{equation}

\begin{figure}
    \centering
    \subfigure[$\hat \beta_{cos}^{FOMFE}$]
    {
        \includegraphics[width=6in]{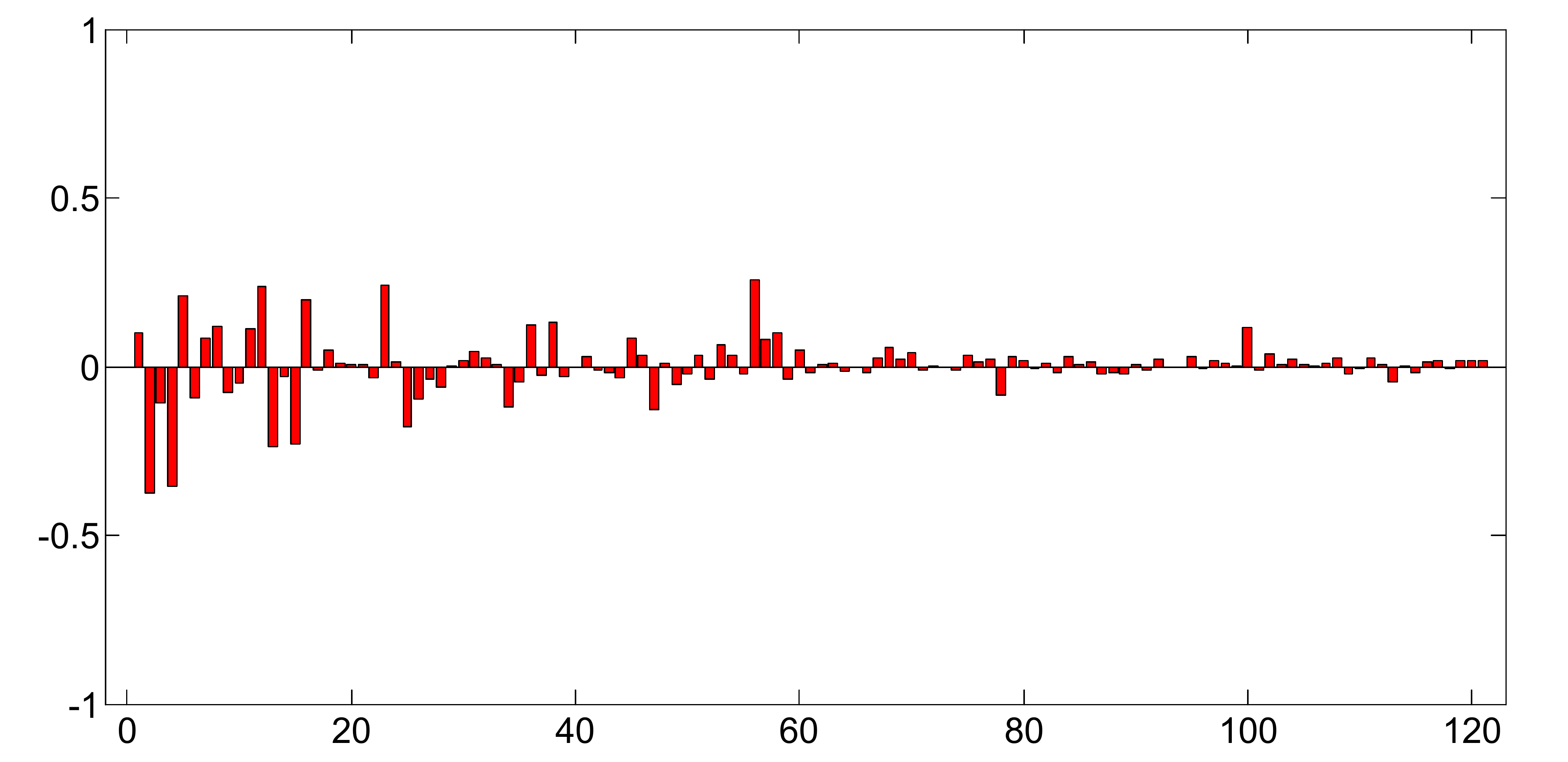}
        \label{classical_and_sparse_FOMFE_coefficient:c_FOMFE_cos}
    }
    \subfigure[$\hat \beta_{sin}^{FOMFE}$]
    {
        \includegraphics[width=6in]{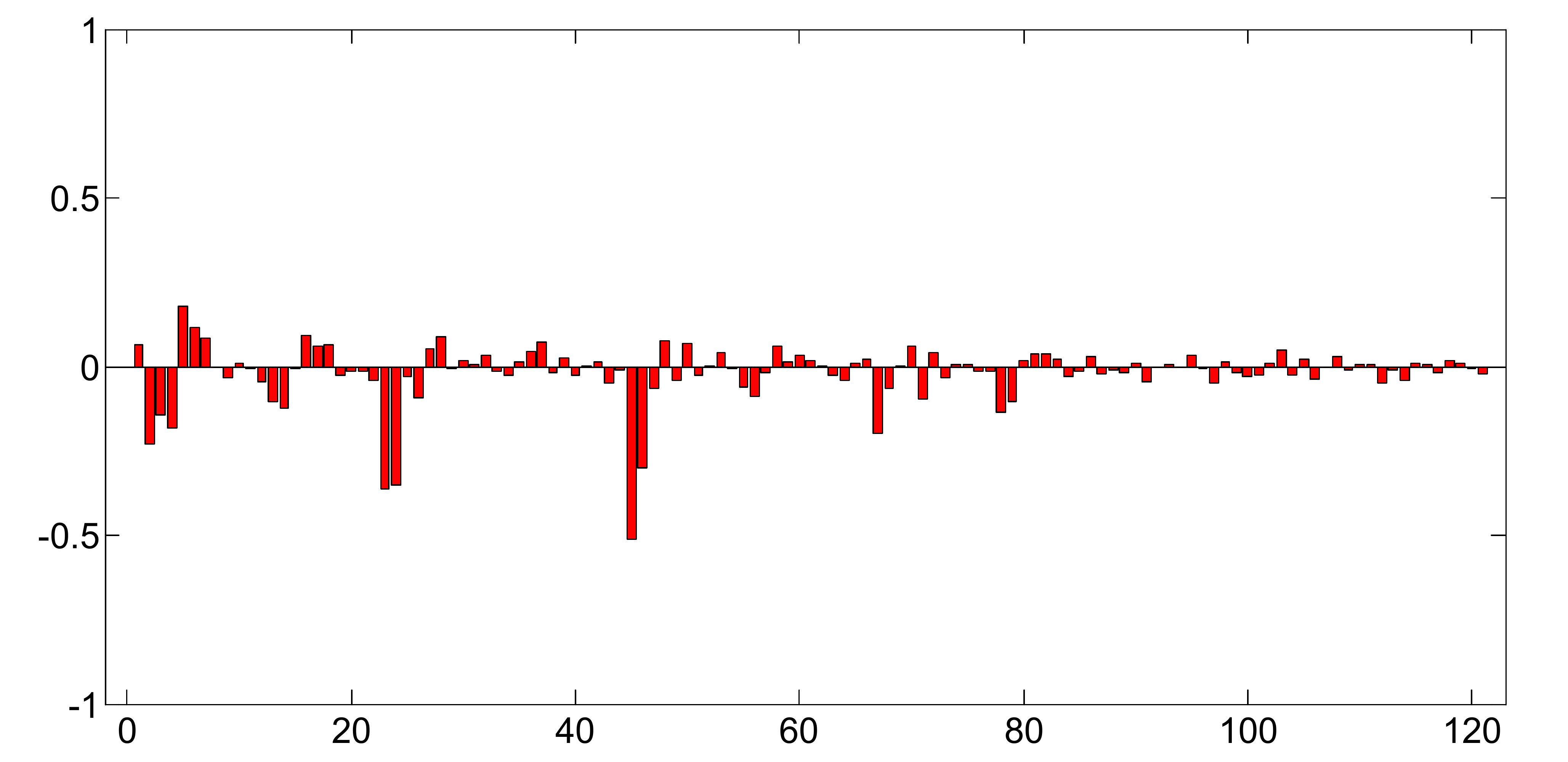}
        \label{classical_and_sparse_FOMFE_coefficient:c_FOMFE_sin}
    }
    \caption{The illustration of the solved classical FOMFE model coefficient vectors: (a) $\hat \beta_{cos}^{FOMFE}$ with $121$ non-zero values and (b) $\hat \beta_{sin}^{FOMFE}$ with $121$ non-zero values.}
    \label{classical_and_sparse_FOMFE_coefficient}
\end{figure}

As an example, the solved classical FOMFE model coefficient vectors $\hat \beta_{cos}^{FOMFE}$ and $\hat \beta_{sin}^{FOMFE}$ are shown in Figure \ref{classical_and_sparse_FOMFE_coefficient:c_FOMFE_cos} and Figure \ref{classical_and_sparse_FOMFE_coefficient:c_FOMFE_sin} respectively.

\subsection{Sparse FOMFE Model (s-FOMFE)}

Based on classical FOMFE model, the $L_1$ norm regulators ${\left\| {{\beta_{cos}}} \right\|_1}$ and ${\left\| {{\beta_{sin}}} \right\|_1}$ are embedded to form the following s-FOMFE model:

\begin{equation}
\hat \beta_{cos}^{s-FOMFE} = \mathop {\arg \min }\limits_{\beta_{cos}} \left\{ {\left\| {b_{cos} - \Phi \beta_{cos}} \right\|_2^2 + \lambda_{cos} {{\left\| \beta_{cos} \right\|}_1}} \right\}
\label{sFOMFE_cos}
\end{equation}

\begin{equation}
\hat \beta_{sin}^{s-FOMFE} = \mathop {\arg \min }\limits_{\beta_{sin}} \left\{ {\left\| {b_{sin} - \Phi \beta_{sin}} \right\|_2^2 + \lambda_{sin} {{\left\| \beta_{sin} \right\|}_1}} \right\}
\label{sFOMFE_sin}
\end{equation}
where ${\left\| {{\beta_{cos}}} \right\|_1}$ and ${\left\| {{\beta_{sin}}} \right\|_1}$ are performed to compel the model coefficient vectors $\beta_{cos}$ and $\beta_{sin}$ to be sparse. The motivation of sparsification for model coefficient vectors is: to represent orientation information $b_{cos}$ and $b_{sin}$ based on the limited number of model coefficients. Actually, the sparse regulators should be ${\left\| {{\beta _{cos}}} \right\|_0}$ and ${\left\| {{\beta _{sin}}} \right\|_0}$, because $L_0$ norm directly constrains the quantity of non-zero values in model coefficient vectors ${\beta_{cos}}$ and ${\beta_{sin}}$ respectively. That is, the model coefficient vectors could be more sparse (more zeros and less non-zero values) if $min({\left\| \cdot \right\|_0})$ could be smaller; otherwise, the model coefficient vectors would become less sparse or even more dense (less zeros and more non-zero values) if $min({\left\| \cdot \right\|_0})$ was larger. However, $L_0$ norm leads to the NP-hard problem. In order to avoid the NP-hard problem, the $L_1$ norm-based regulators ${\left\| {{\beta_{cos}}} \right\|_1}$ and ${\left\| {{\beta_{sin}}} \right\|_1}$ are applied to approximate the corresponding $L_0$ norm-based regulators ${\left\| {{\beta _{cos}}} \right\|_0}$ and ${\left\| {{\beta _{sin}}} \right\|_0}$ respectively.

\begin{figure}
    \centering
    \subfigure[$\hat \beta_{cos}^{s-FOMFE}$]
    {
        \includegraphics[width=6in]{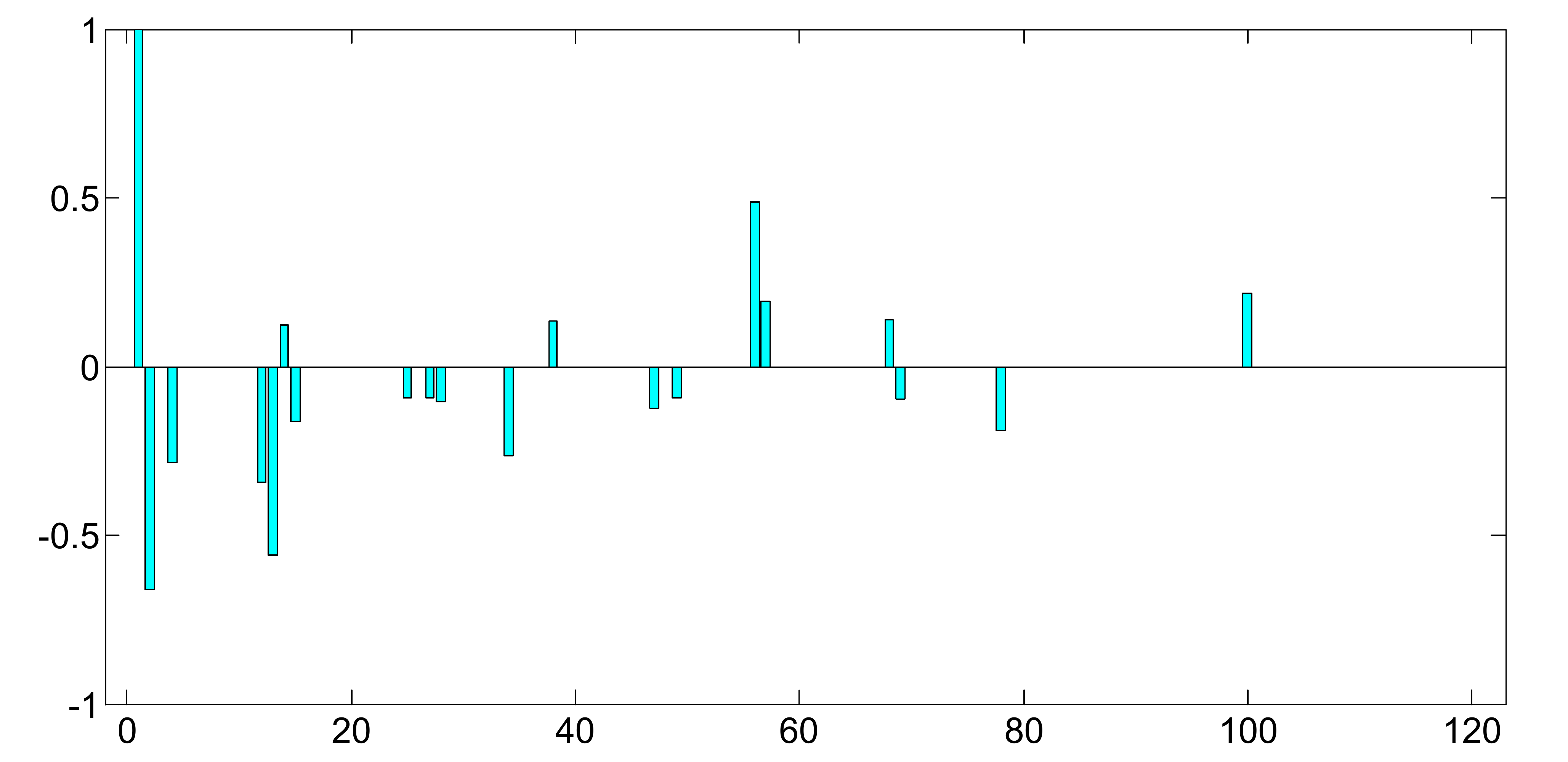}
        \label{classical_and_sparse_FOMFE_coefficient:s_FOMFE_cos}
    }
    \subfigure[$\hat \beta_{sin}^{s-FOMFE}$]
    {
        \includegraphics[width=6in]{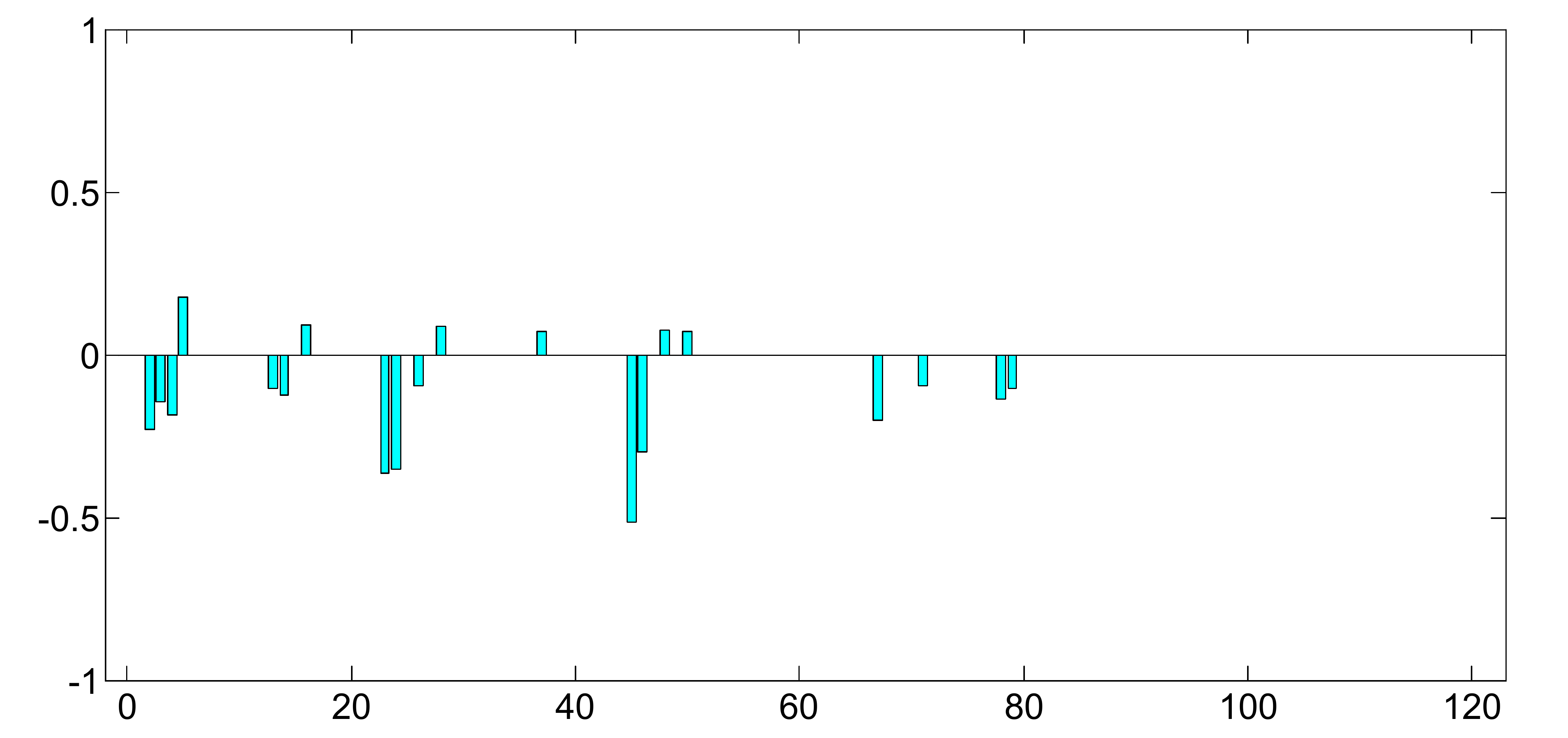}
        \label{classical_and_sparse_FOMFE_coefficient:s_FOMFE_sin}
    }
    \caption{The illustration of the solved sparse FOMFE model coefficient vectors: (a) $\hat \beta_{cos}^{s-FOMFE}$ with $20$ non-zero values and (b) $\hat \beta_{sin}^{s-FOMFE}$ with $20$ non-zero values.}
    \label{classical_and_sparse_FOMFE_coefficient}
\end{figure}

For solving s-FOMFE model coefficient vectors $\beta_{cos}$ and $\beta_{sin}$ in Eqs. (\ref{sFOMFE_cos}) and (\ref{sFOMFE_sin}), orthogonal matching pursuit (OMP) \cite{Tropp2004} is employed. As an example, the solved $\hat \beta_{cos}^{s-FOMFE}$ and $\hat \beta_{sin}^{s-FOMFE}$ are shown in Figure \ref{classical_and_sparse_FOMFE_coefficient:s_FOMFE_cos} and Figure \ref{classical_and_sparse_FOMFE_coefficient:s_FOMFE_sin} respectively.

\subsection{Compressed Sparse FOMFE Model (cs-FOMFE)}

The introduced s-FOMFE model can be further modified by exploiting the compressed sensing (CS) theory. Incorporating with CS, the compressed sparse FOMFE model (cs-FOMFE) is mathematically formulated as follows:

\begin{equation}
\hat \beta_{cos}^{cs-FOMFE} = \mathop {\arg \min }\limits_{\beta_{cos}} \left\{ {\left\| {g_{cos} - \Psi \Phi \beta_{cos}} \right\|_2^2 + \lambda_{cos} {{\left\| \beta_{cos} \right\|}_1}} \right\}
\label{csFOMFE_cos}
\end{equation}

\begin{equation}
\hat \beta_{sin}^{cs-FOMFE} = \mathop {\arg \min }\limits_{\beta_{sin}} \left\{ {\left\| {g_{sin} - \Psi \Phi \beta_{sin}} \right\|_2^2 + \lambda_{sin} {{\left\| \beta_{sin} \right\|}_1}} \right\}
\label{csFOMFE_sin}
\end{equation}
where $\Psi$ ($m \times n$) is a compressed sensing (measuring) matrix. Besides ${g_{cos}} = \Psi {b_{cos}}$ and ${g_{sin}} = \Psi {b_{sin}}$ are compressed measurements based on $\Psi$.

\begin{figure}
\centerline{\includegraphics[height=4.0in]{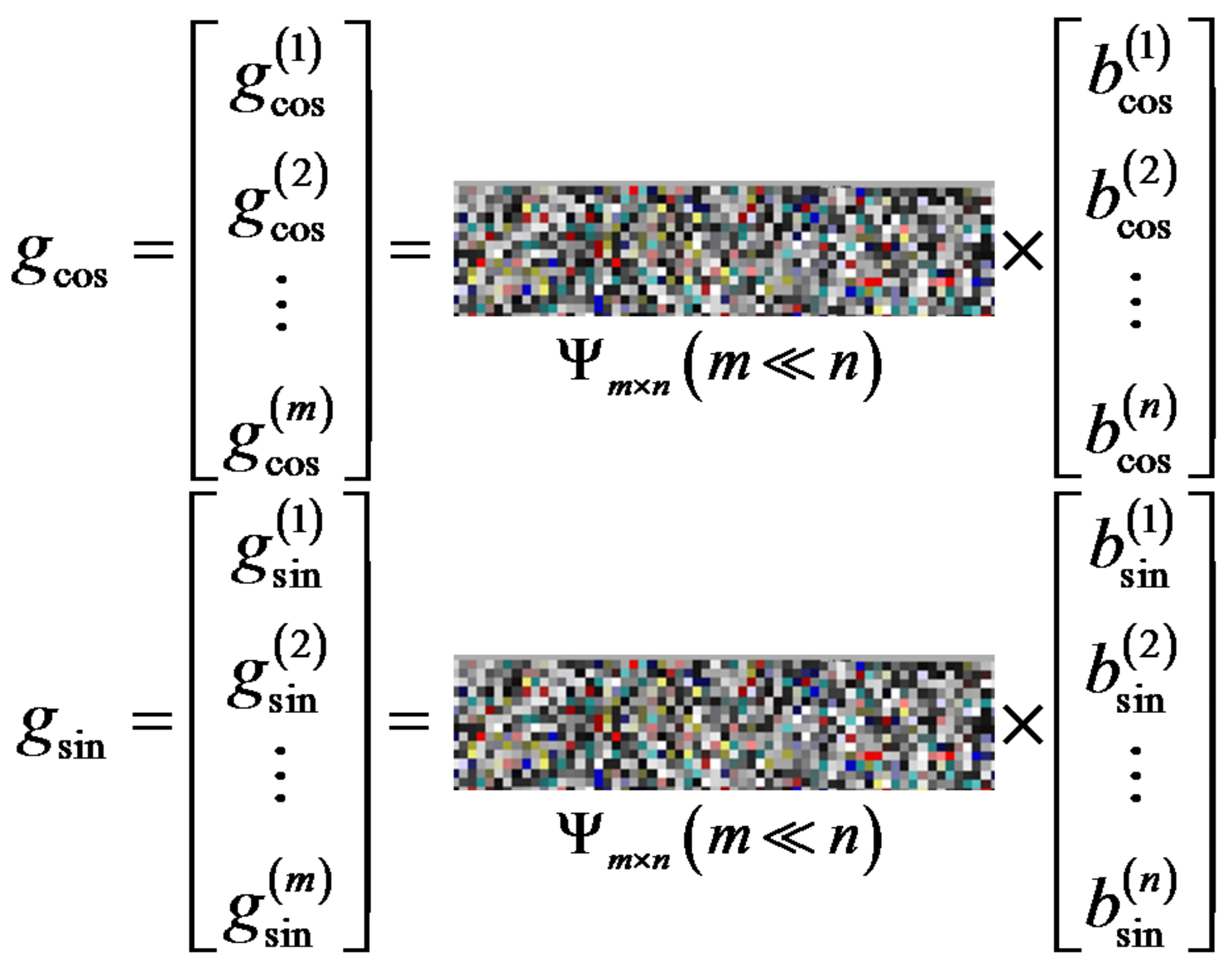}}
\caption{An example of random Gaussian matrix-based compressed sensing (measuring) procedure.}
\label{random_sensing_procedure}
\end{figure}

As theoretically guaranteed by CS theory, even the measurements for fingerprint OF ($m$: the rows of $\Psi$) could be much smaller than the vector dimension of $b_{cos}$ or $b_{sin}$ ($n$: the columns of $\Psi$ and also the vector dimension of $b_{cos}$ or $b_{sin}$), the approximate recovery of original fingerprint OF is still able to be achieved relying on $L_1$ norm minimization. Such fingerprint OF reconstruction procedure is conducted within the undersampling and sparse signal representation framework. Particularly, in order to preserve the geometric structure of the original fingerprint OF in compressed measurement manifold (undersampled space), the restricted isometry property (RIP) needs to be ensured. To be explicit, the signal representation should be performed via a set of locally coherent basis, while the compressed measuring has to be implemented by globally incoherent sensing matrix. As theoretically proven and practically verified, the random Gaussian matrix is indeed incoherent in undersampled space. Besides, the available measurements $m$ needs to satisfy the following condition:

\begin{equation}
m \ge C \cdot S \cdot \log \left( {{n \mathord{\left/
 {\vphantom {n S}} \right.
 \kern-\nulldelimiterspace} S}} \right)
\label{measurement_condition}
\end{equation}
where $C$ is a constant and usually $C=10$ is chosen in practice. $S$ denotes the sparseness constraint for OF model coefficient vectors $\beta_{cos}$ and $\beta_{sin}$ (for example, $S=20$ regulates that the non-zero values in both vectors $\beta_{cos}$ and $\beta_{sin}$ should be less than or equal to 20). $n$ is the dimension of original observation vectors $b_{cos}$ and $b_{sin}$. As an example, a random Gaussian matrix-based compressed sensing (measuring) procedure is illustrated in Figure \ref{random_sensing_procedure}.

\begin{figure}
    \centering
    \subfigure[$\hat \beta_{cos}^{cs-FOMFE}$]
    {
        \includegraphics[width=6in]{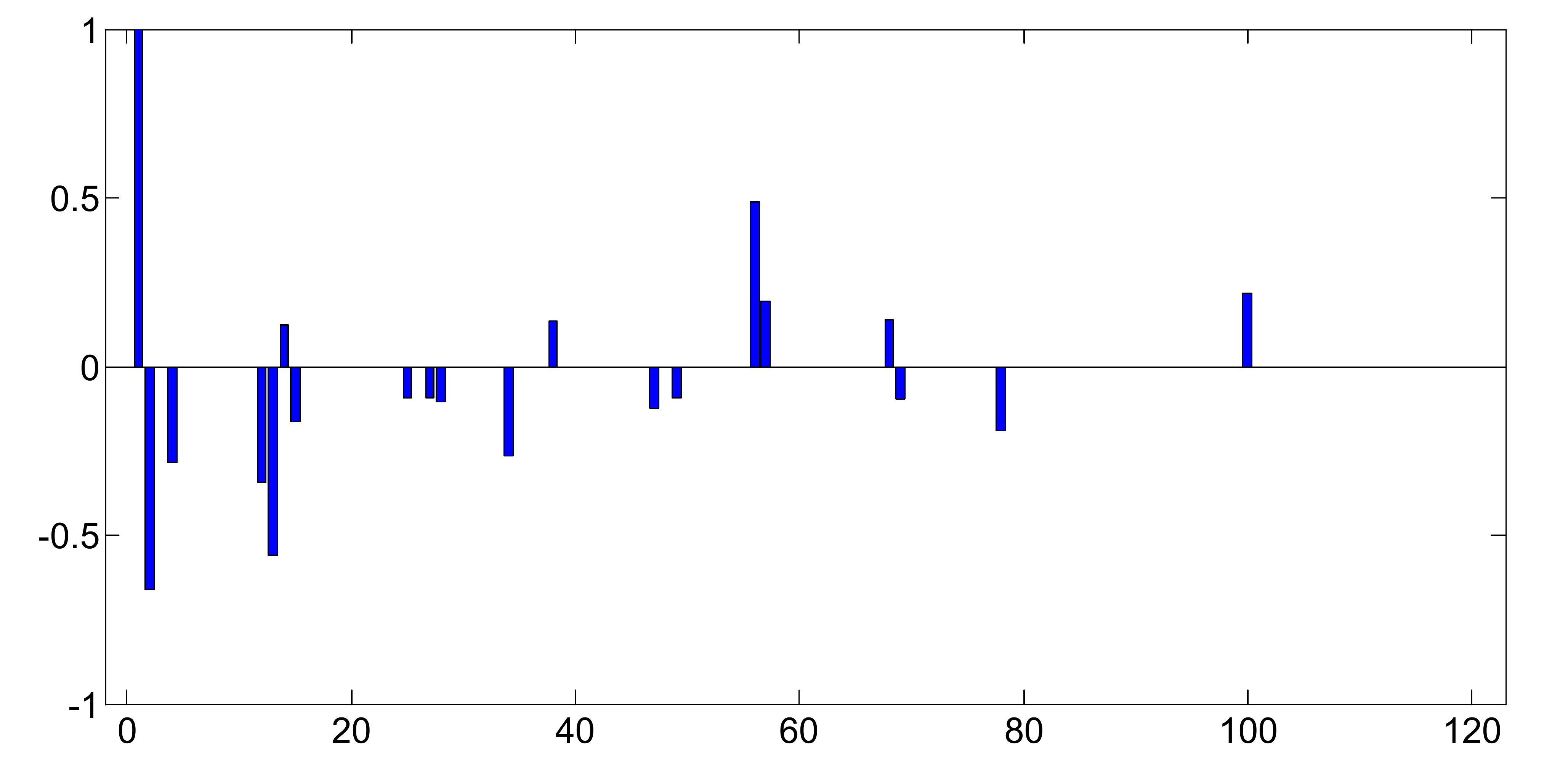}
        \label{classical_and_compressed_sparse_FOMFE_coefficient:cs_FOMFE_cos}
    }
    \subfigure[$\hat \beta_{sin}^{cs-FOMFE}$]
    {
        \includegraphics[width=6in]{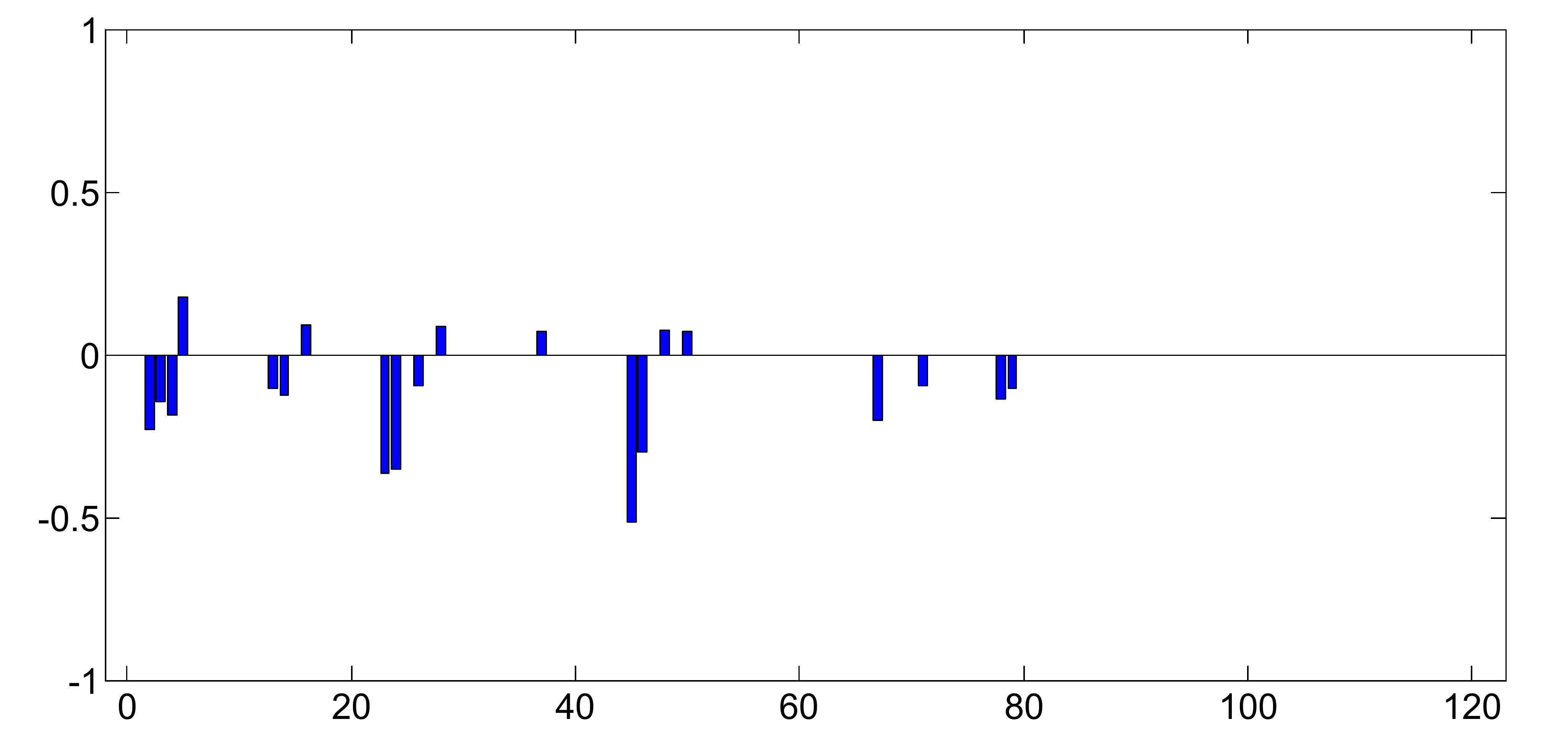}
        \label{classical_and_compressed_sparse_FOMFE_coefficient:cs_FOMFE_sin}
    }
    \caption{The illustration of the solved compressed sparse FOMFE model coefficient vectors: (a) $\hat \beta_{cos}^{cs-FOMFE}$ with $20$ non-zero values and (b) $\hat \beta_{sin}^{cs-FOMFE}$ with $20$ non-zero values.}
    \label{classical_and_compressed_sparse_FOMFE_coefficient}
\end{figure}

For solving cs-FOMFE model coefficient vectors $\beta_{cos}$ and $\beta_{sin}$ in Eqs. (\ref{csFOMFE_cos}) and (\ref{csFOMFE_sin}), OMP is still utilized. As an example, the solved $\hat \beta_{cos}^{cs-FOMFE}$ and $\hat \beta_{sin}^{cs-FOMFE}$ are shown in Figure \ref{classical_and_compressed_sparse_FOMFE_coefficient:cs_FOMFE_cos} and Figure \ref{classical_and_compressed_sparse_FOMFE_coefficient:cs_FOMFE_sin} respectively.

\section{Experiment}

In this section, the proposed FOMFE models: s-FOMFE and cs-FOMFE are tested by the following three experiments: (i) OF modeling based on plain fingerprint image; (ii) OF modeling based on rolled fingerprint image; and (iii) OF modeling based on latent fingerprint image.

\subsection{Data Preparation}

\begin{figure}
    \centering
    \subfigure[]
    {
        \includegraphics[height=3in]{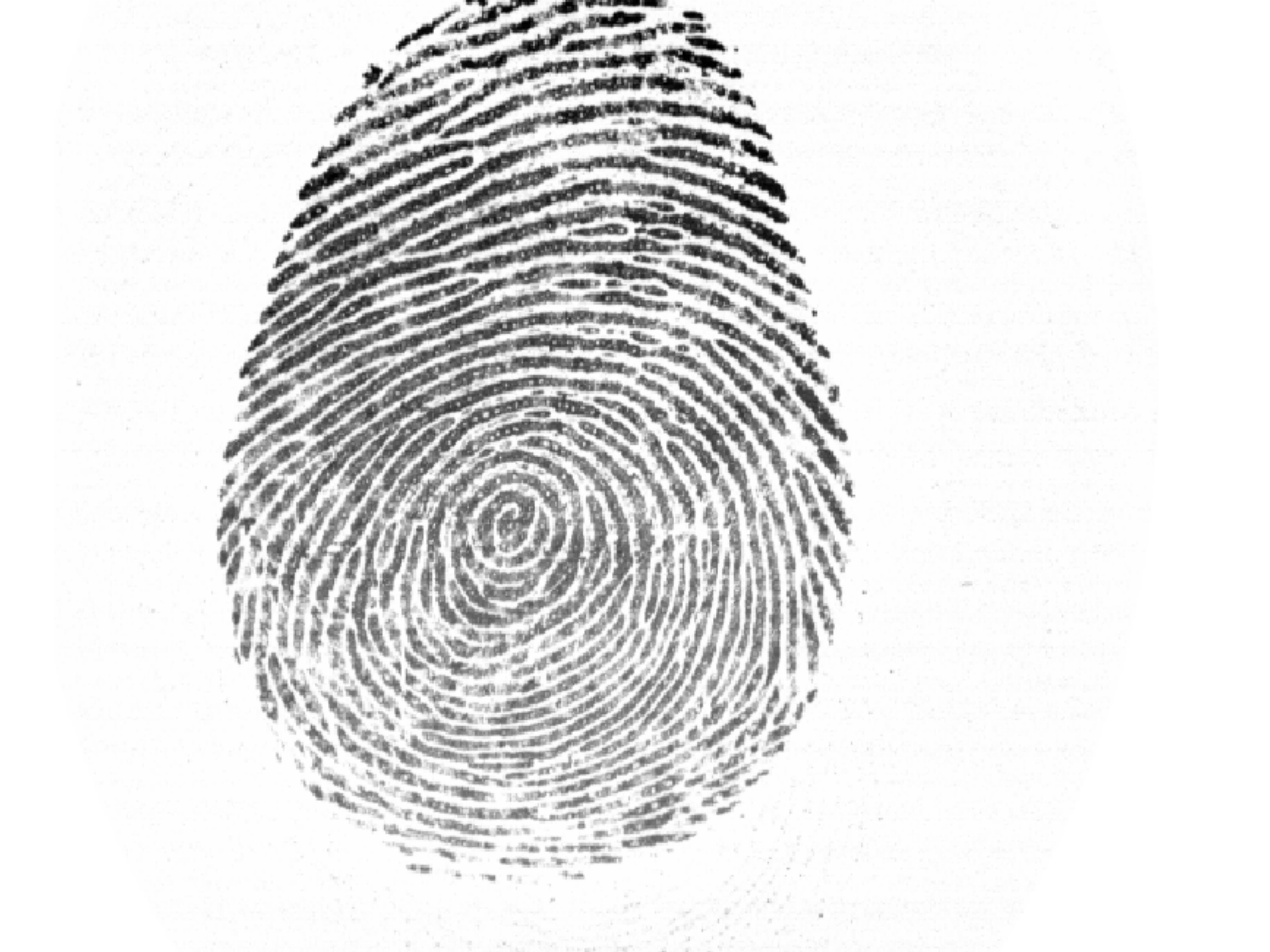}
        \label{plain_rolled_latent_dataset:plain}
    }
    \\
    \subfigure[]
    {
        \includegraphics[width=3in]{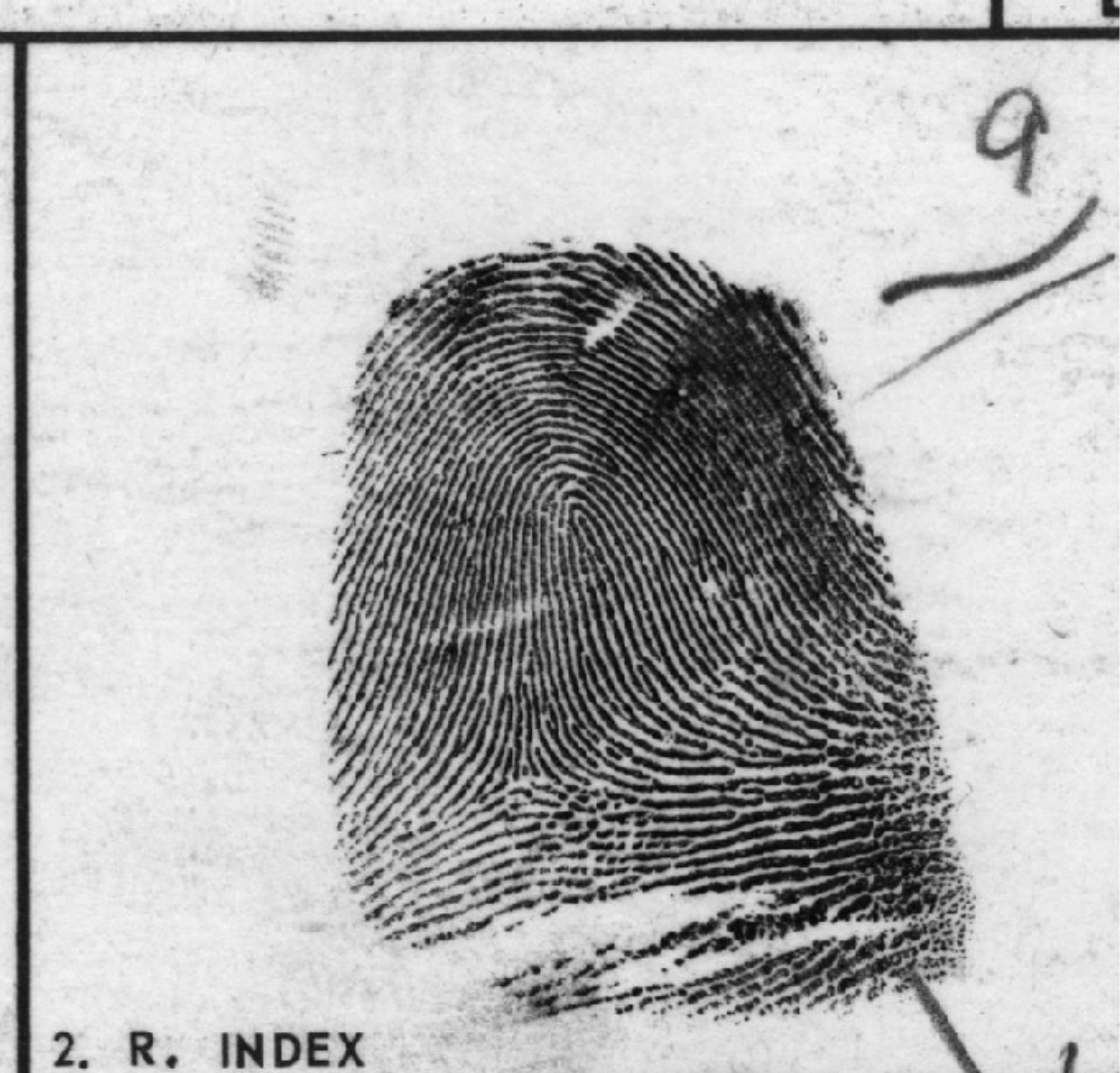}
        \label{plain_rolled_latent_dataset:rolled}
    }
    \subfigure[]
    {
        \includegraphics[width=3in]{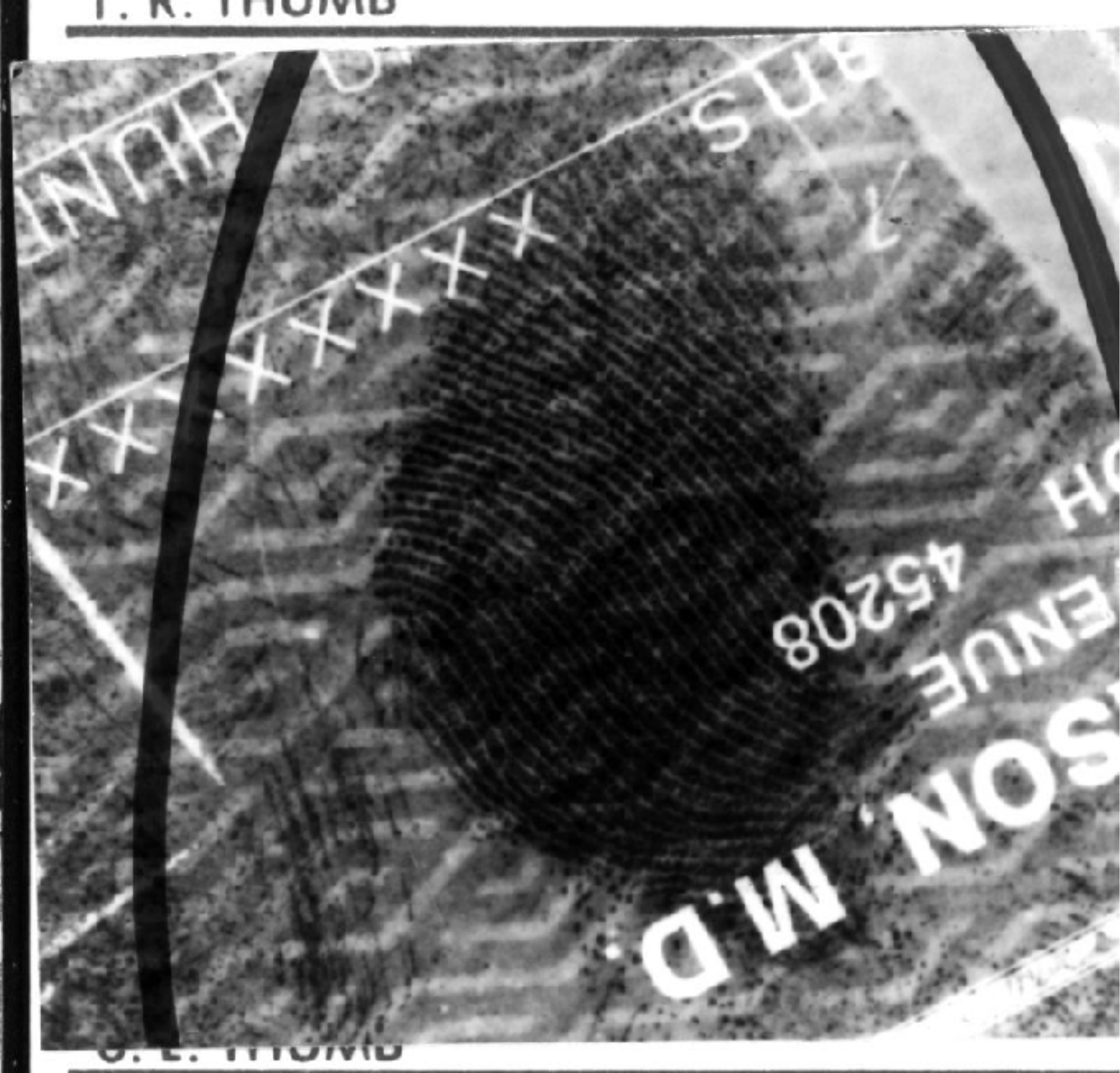}
        \label{plain_rolled_latent_dataset:latent}
    }
    \caption{The used fingerprint images for OF modeling experiments: (a) the plain print image; (b) the rolled print image and (c) the latent print image.}
    \label{plain_rolled_latent_dataset}
\end{figure}

All the experiments are conducted on a private plain fingerprint database (ADFA 2D) and a public available latent fingerprint database (NIST SD27). ADFA 2D fingerprint database consists of $600$ plain print images, which are captured by a commercial pressing-down scanner. NIST SD27 fingerprint database includes $258$ latent print images and their corresponding rolled print images, which are accessible in the public domain. Such fingerprint images used for OF modeling experiments are shown in Figure \ref{plain_rolled_latent_dataset}.

Because no ground truth exists for the evaluation of constructed fingerprint OF, the objective error measurement cannot be easily created and adopted \cite{Zhou2004} \cite{Wang2007}, it is difficult to judge the quality of constructed OF in a quantitative way. Instead, the quality of constructed OF has to be assessed by means of visual inspection. For the following three experiments, the OF modeling results obtained by three different FOMFE-based models: classical FOMFE, s-FOMFE and cs-FOMFE are illustrated and visually compared.

\subsection{OF Modeling Experiment on Plain Fingerprint Image}

\begin{figure}
    \centering
    \subfigure[]
    {
        \includegraphics[width=3in]{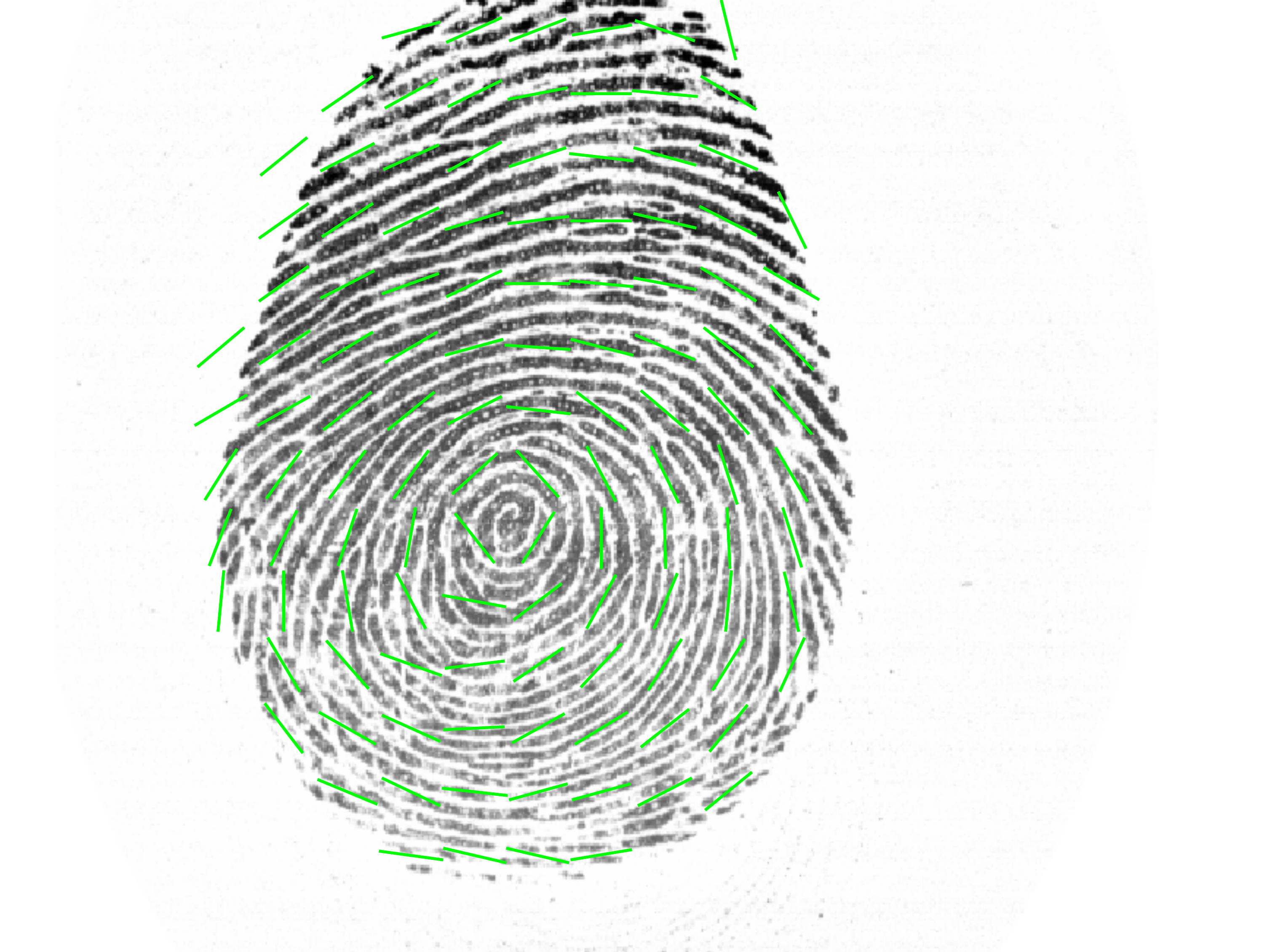}
        \label{plain_OF_modeling:plain_coarse_OF}
    }
    \subfigure[]
    {
        \includegraphics[width=3in]{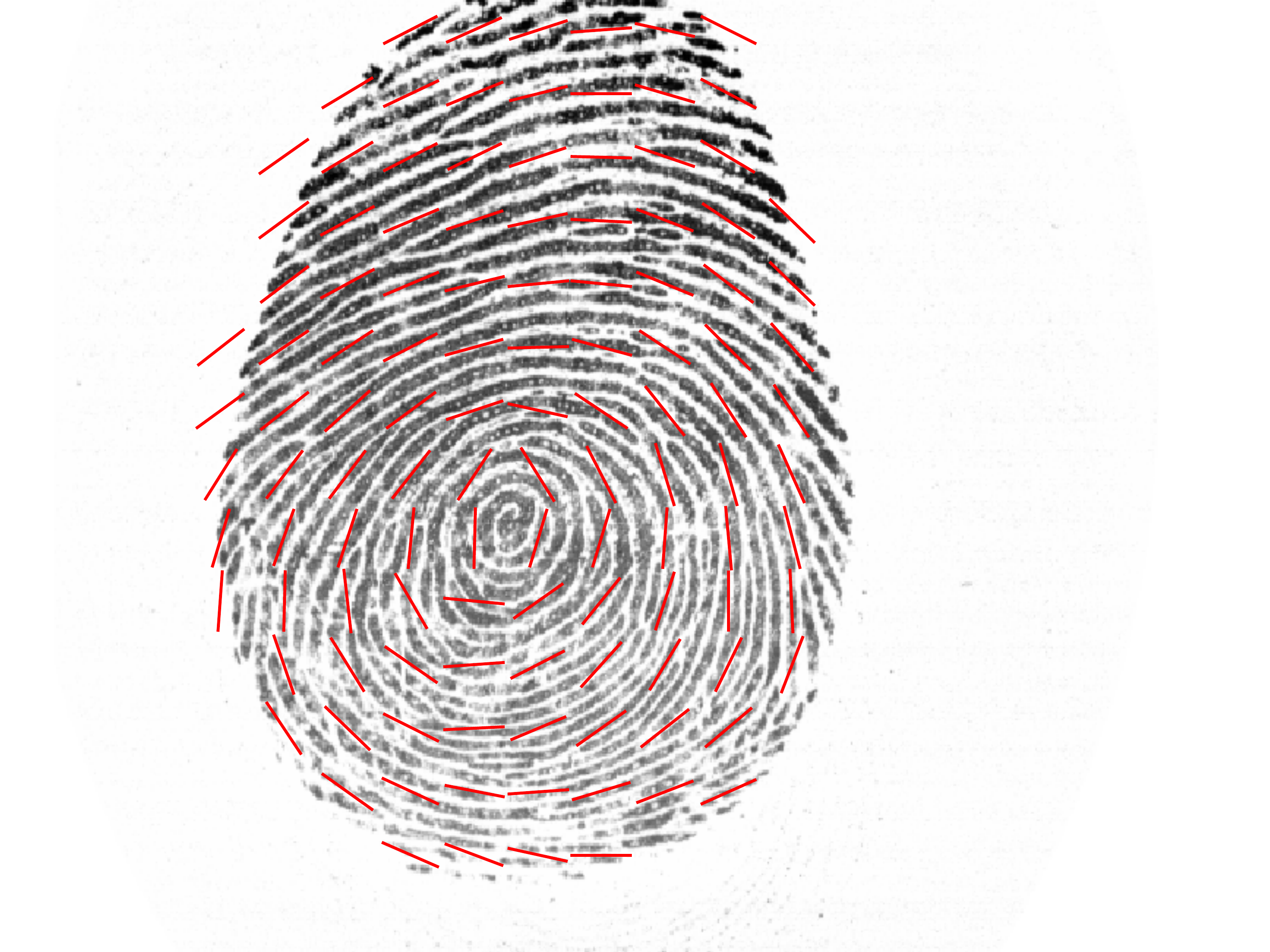}
        \label{plain_OF_modeling:plain_classical_FOMFE_OF}
    }
    \subfigure[]
    {
        \includegraphics[width=3in]{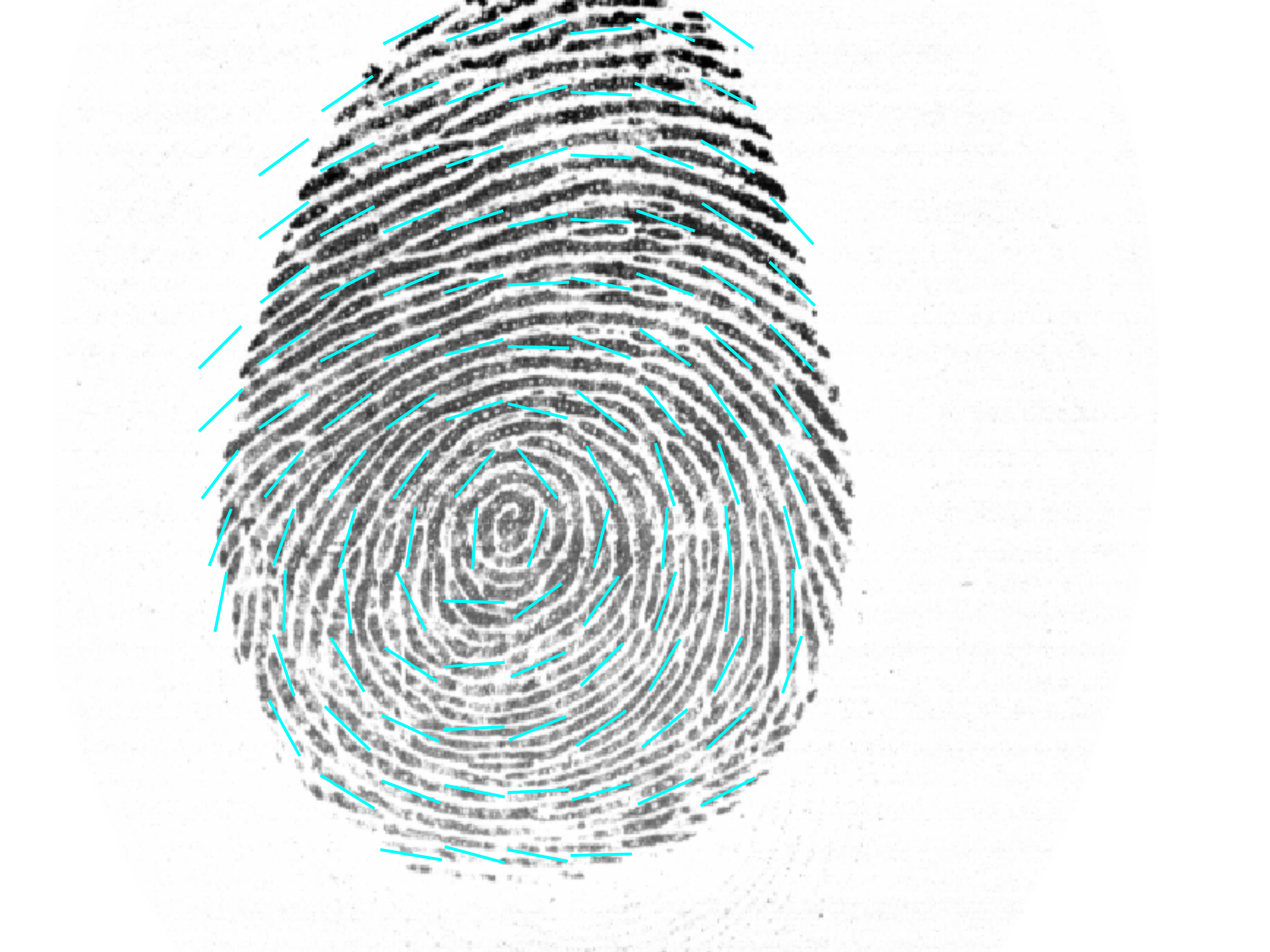}
        \label{plain_OF_modeling:plain_sparse_FOMFE_OF}
    }
    \subfigure[]
    {
        \includegraphics[width=3in]{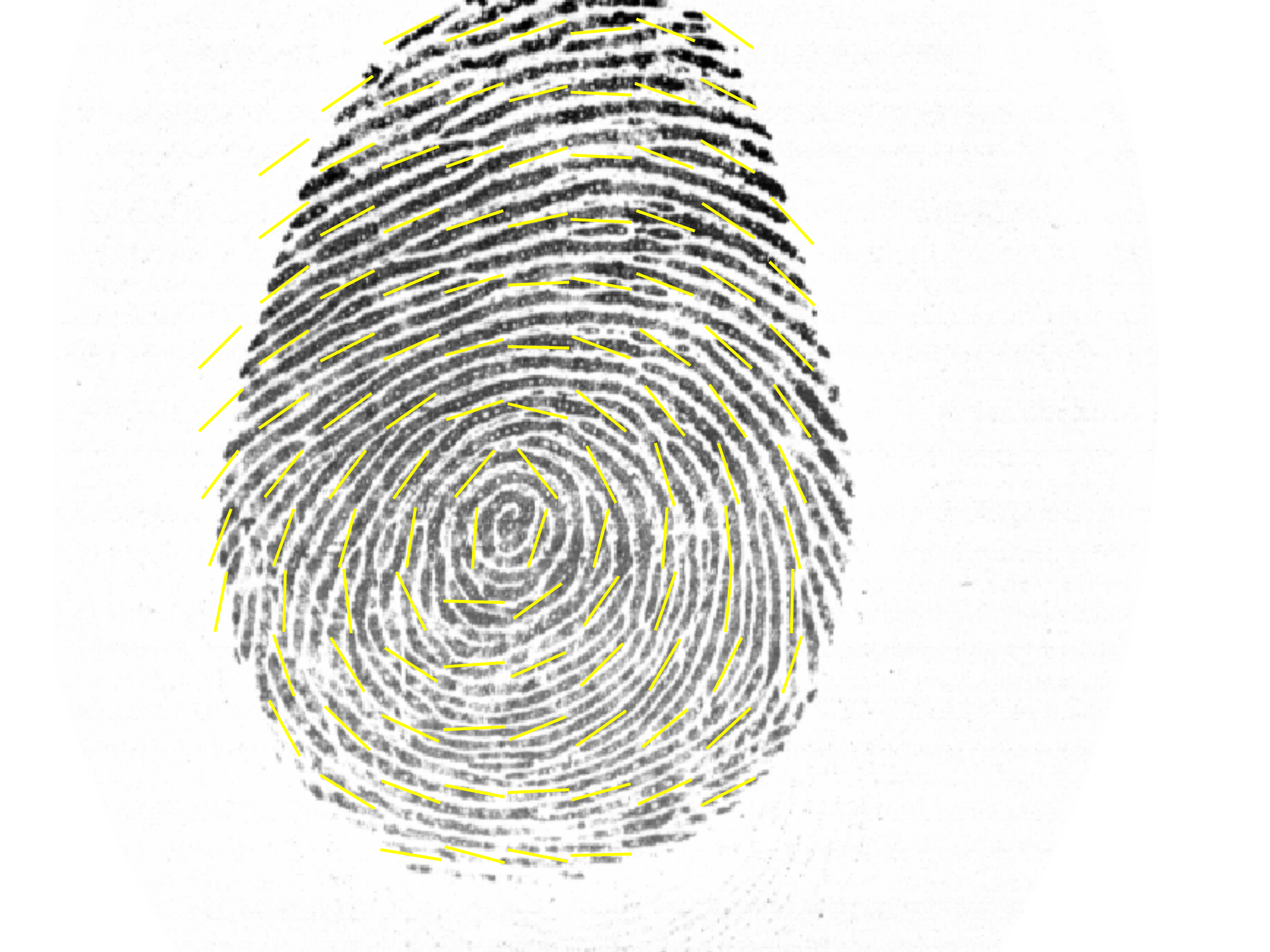}
        \label{plain_OF_modeling:plain_compressed_sparse_FOMFE_OF}
    }
    \caption{The illustration of coarse OF, classical FOMFE OF, s-FOMFE OF and cs-FOMFE OF for plain print image respectively: (a) coarse OF estimated by \cite{Maltoni2009}; (b) classical FOMFE OF; (c) s-FOMFE OF and (d) cs-FOMFE OF.}
    \label{plain_OF_modeling}
\end{figure}

Given the coarse OF estimated by \cite{Maltoni2009} as the models input (shown in Figure \ref{plain_OF_modeling:plain_coarse_OF}), the modeling parameters are tuned as follows: image block size $w = 16$, FOMFE basis order $k = 5$, and sparseness constraint $S = 20$. Consequently, the constructed OFs via classical FOMFE model, s-FOMFE model, and cs-FOMFE model are shown in Figure \ref{plain_OF_modeling:plain_classical_FOMFE_OF}, Figure \ref{plain_OF_modeling:plain_sparse_FOMFE_OF} and Figure \ref{plain_OF_modeling:plain_compressed_sparse_FOMFE_OF} respectively. For OF modeling results obtained by classical FOMFE, s-FOMFE and cs-FOMFE, the orientation trend of ridge structure seems quite similar. However, in contrast to classical FOMFE, the model coefficient vectors for s-FOMFE and cs-FOMFE are significantly sparse. To be explicit, only 20 non-zero values exists for model coefficient vectors $\beta_{cos}$ and $\beta_{sin}$ whose vector dimension is ${121 = \left( {2k + 1} \right)^2}$ when $k = 5$.

\subsection{OF Modeling Experiment on Rolled Fingerprint Image}

\begin{figure}
    \centering
    \subfigure[]
    {
        \includegraphics[width=3in]{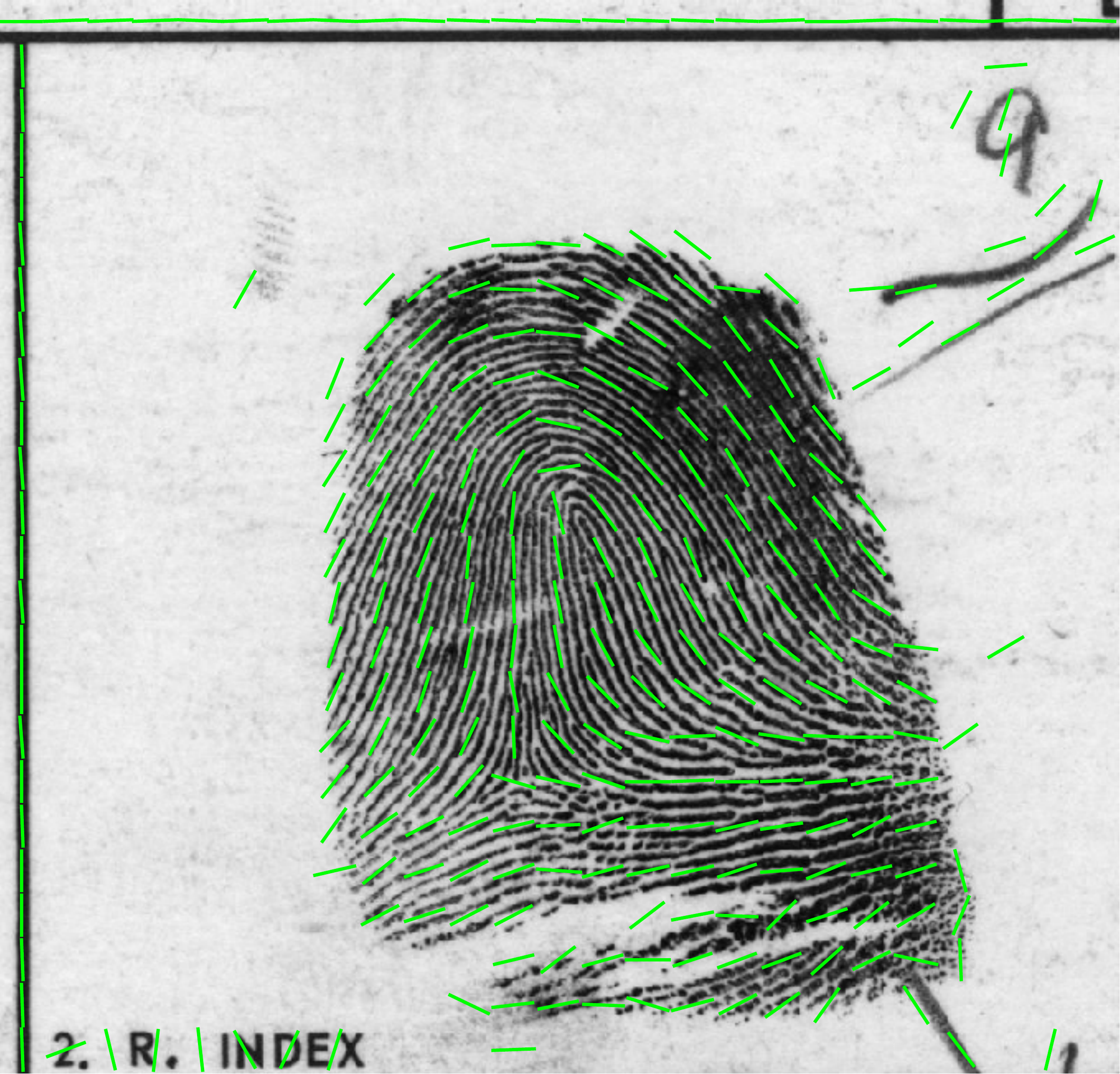}
        \label{rolled_OF_modeling:rolled_coarse_OF}
    }
    \subfigure[]
    {
        \includegraphics[width=3in]{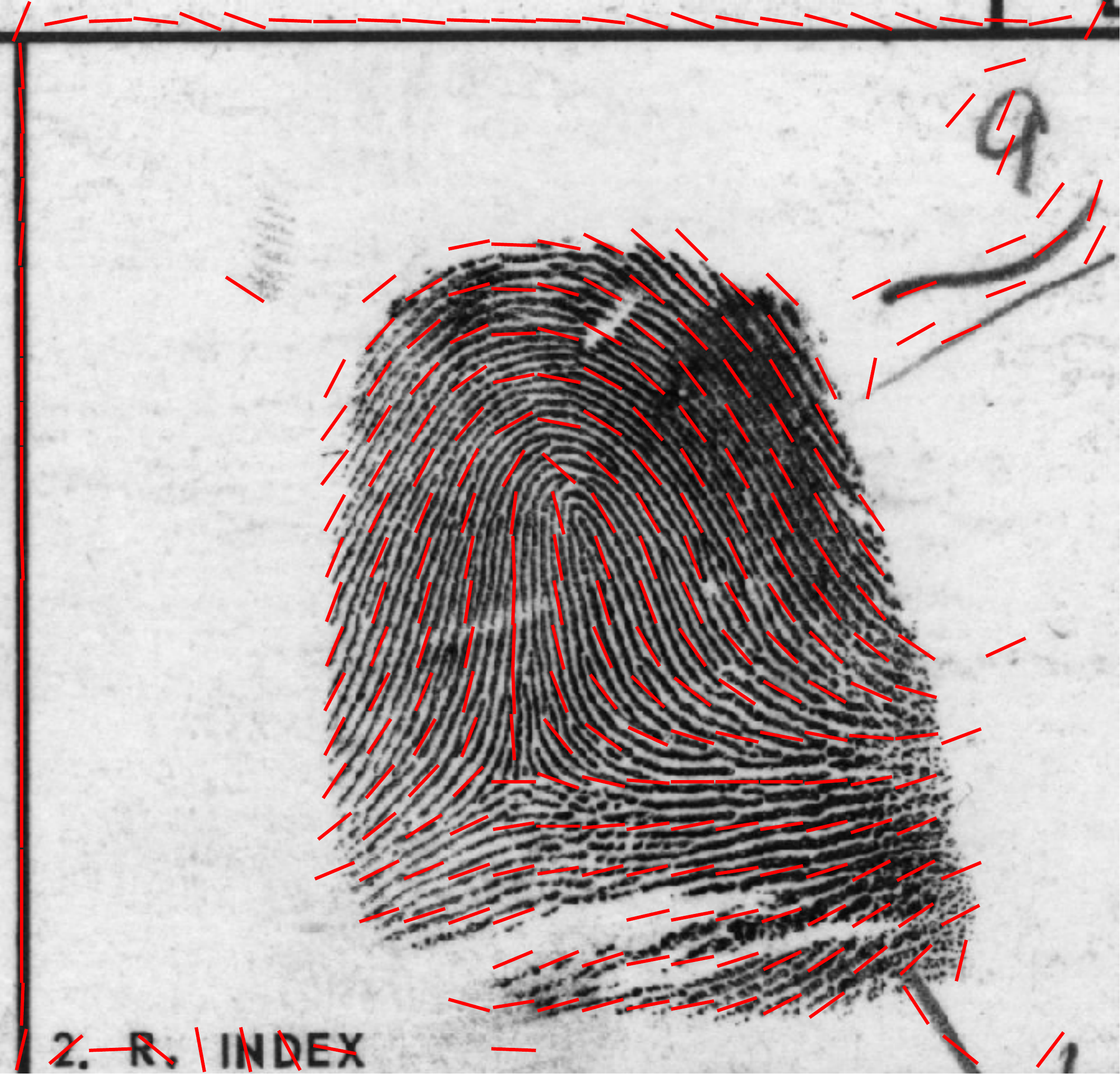}
        \label{rolled_OF_modeling:rolled_classical_FOMFE_OF}
    }
    \subfigure[]
    {
        \includegraphics[width=3in]{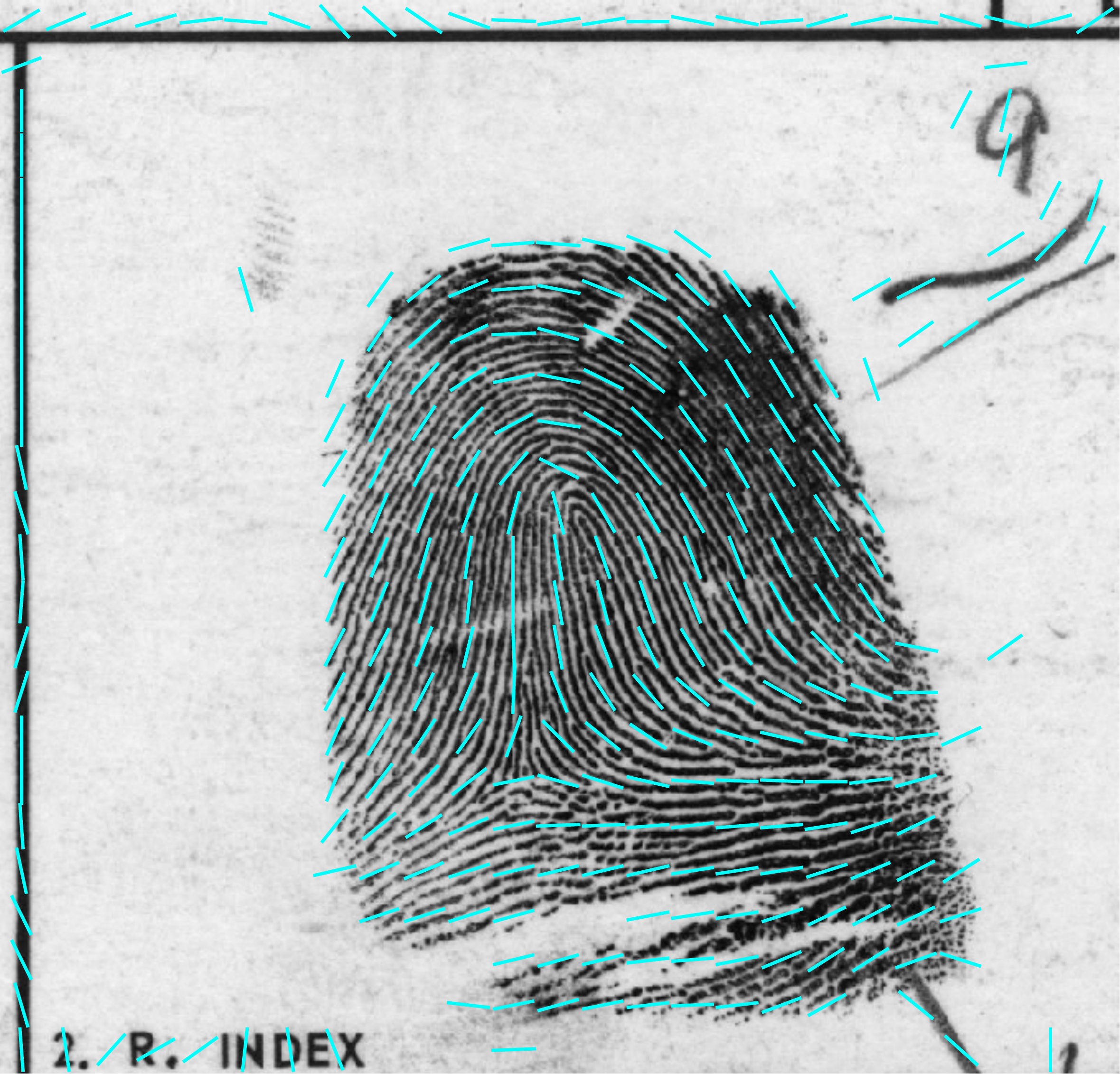}
        \label{rolled_OF_modeling:rolled_sparse_FOMFE_OF}
    }
    \subfigure[]
    {
        \includegraphics[width=3in]{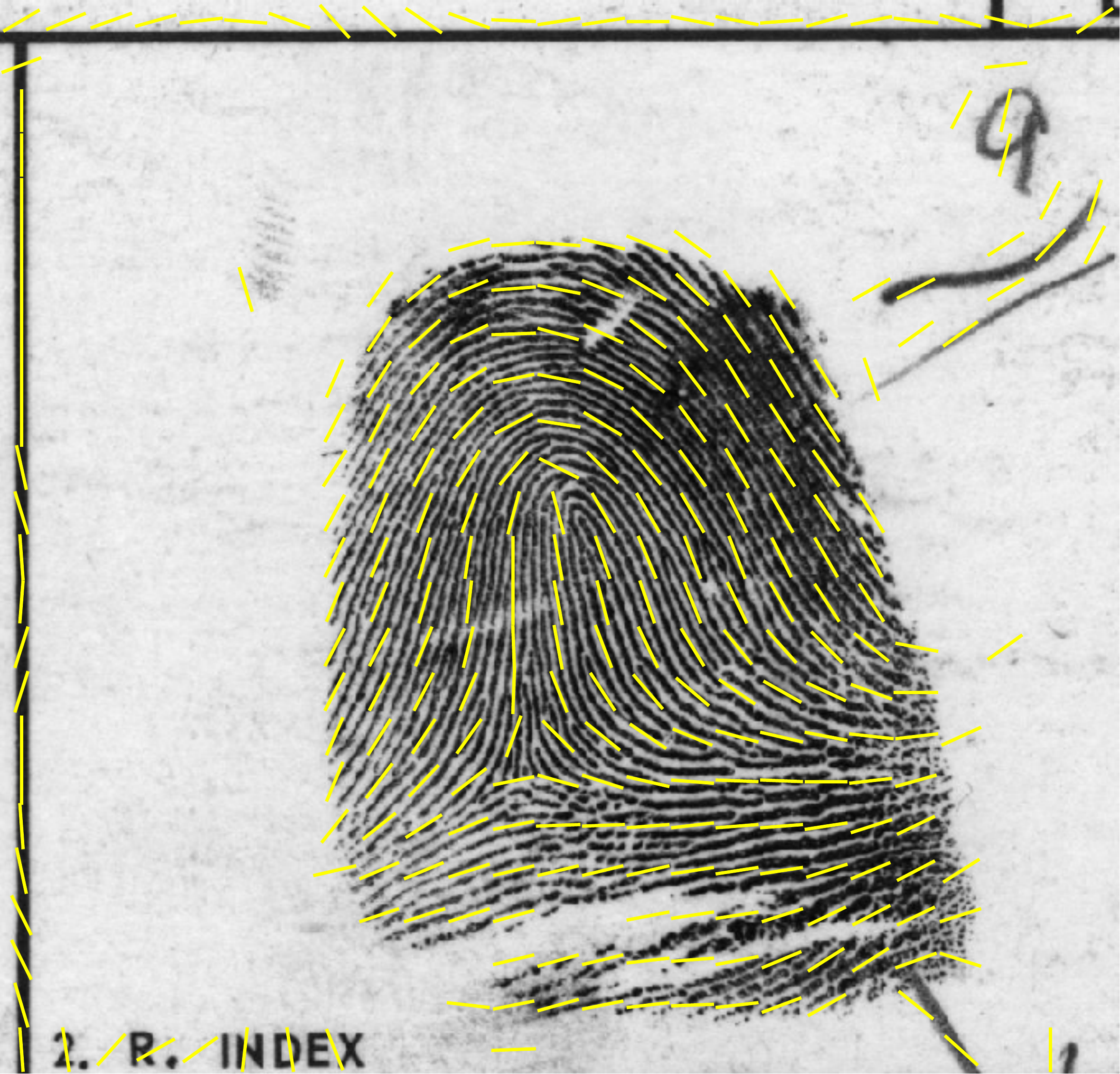}
        \label{rolled_OF_modeling:rolled_compressed_sparse_FOMFE_OF}
    }
    \caption{The illustration of coarse OF, classical FOMFE OF, s-FOMFE OF and cs-FOMFE OF for rolled print image respectively: (a) coarse OF estimated by \cite{Maltoni2009}; (b) classical FOMFE OF; (c) s-FOMFE OF and (d) cs-FOMFE OF.}
    \label{rolled_OF_modeling}
\end{figure}

Similar to the OF modeling experiment on plain print image, the same coarse OF estimation method and modeling parameters are applied to the rolled print image. Accordingly, the constructed OFs via three different FOMFE models are illustrated in Figure \ref{rolled_OF_modeling}. The visual comparison among three different FOMFE models demonstrates that the very similar OF modeling results can be attained and particularly the resultant $\beta_{cos}$ and $\beta_{sin}$ regarding to s-FOMFE and cs-FOMFE are more sparse rather than the ones of classical FOMFE.

\subsection{OF Modeling Experiment on Latent Fingerprint Image}

\begin{figure}
    \centering
    \subfigure[]
    {
        \includegraphics[width=3in]{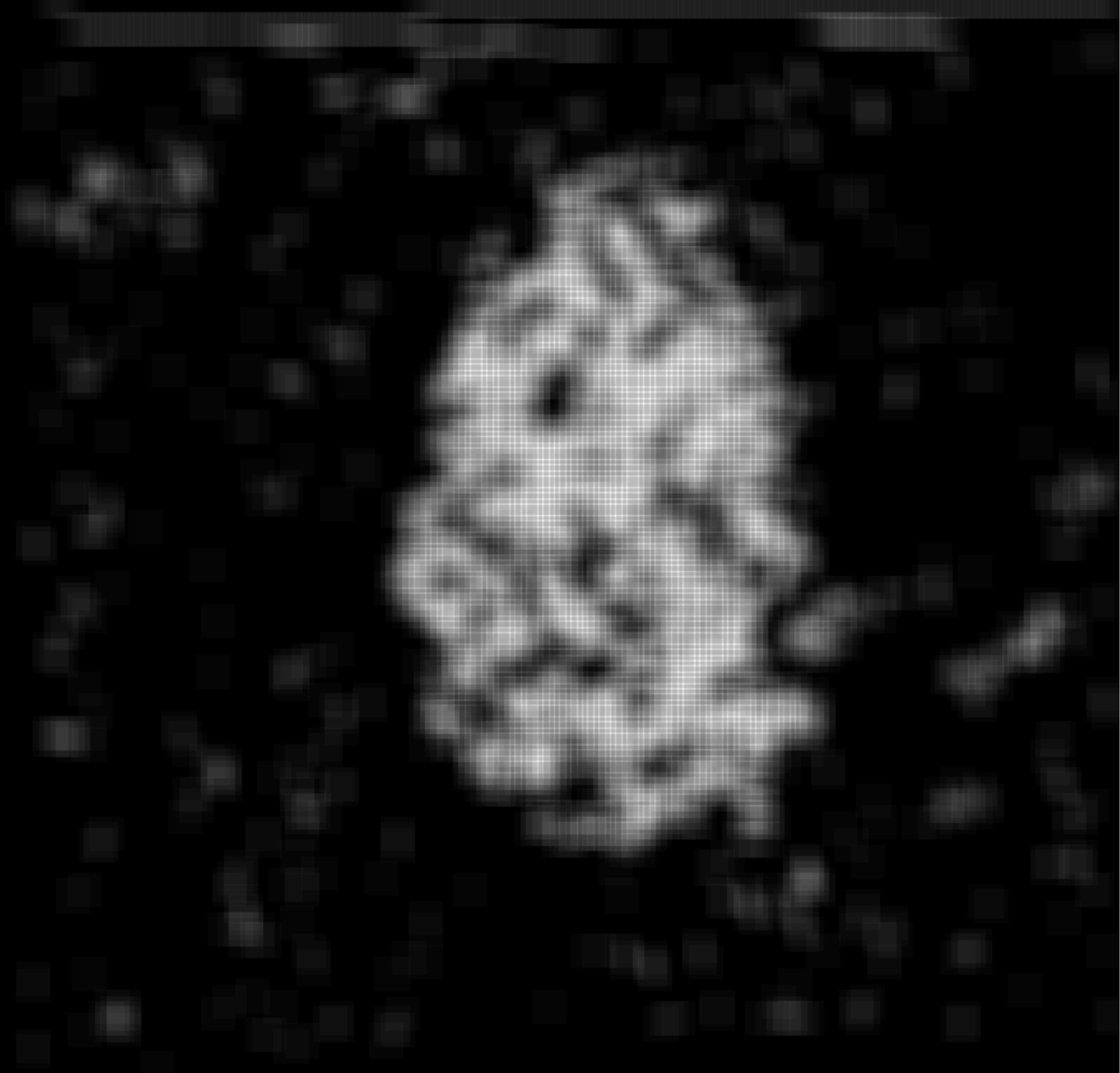}
        \label{high_quality_map:high_quality_map}
    }
    \subfigure[]
    {
        \includegraphics[width=3in]{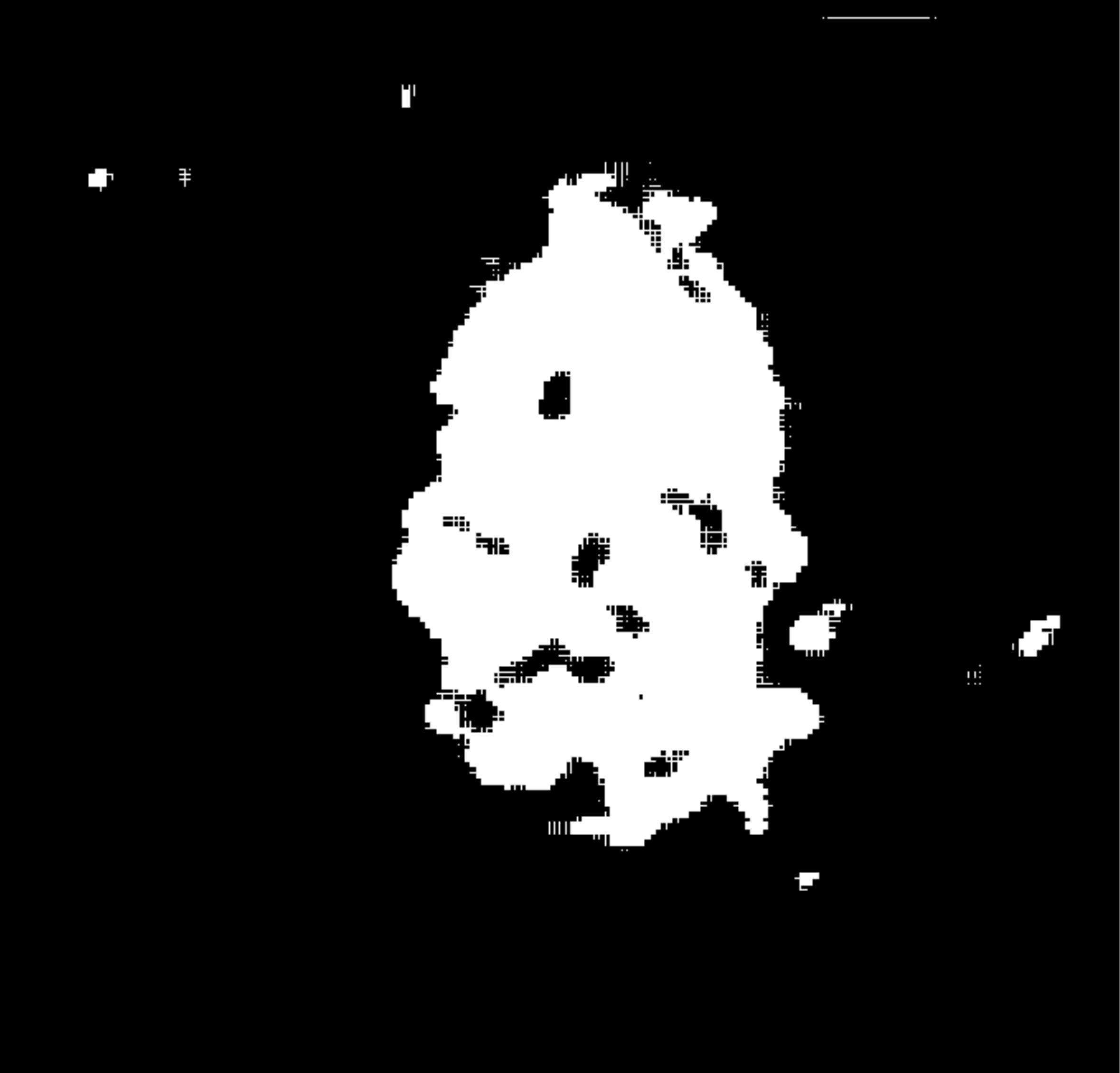}
        \label{high_quality_map:high_quality_mask}
    }
    \caption{The illustration of salient ridge segments and the corresponding spatial mask: (a) salient ridge map and (b) the corresponding spatial mask.}
    \label{high_quality_map}
\end{figure}

In contrast to the plain and rolled fingerprint images, the print in latent images is partial, overlapped with other image content and blurred. Due to such intricate circumstance, the latent print image is generally in poor quality so that the OF modeling for latent fingerprint is challenging. In order to ensure the reliability of OF modeling, the latent fingerprint image needs to be processed prior to the OF modeling stage. Therefore, a dictionary learning-based algorithm is exploited to label the image regions with salient ridge structure \cite{Xu2014}. The salient ridge map and the corresponding spatial mask are shown in Figure \ref{high_quality_map}.

\begin{figure}
    \centering
    \subfigure[]
    {
        \includegraphics[width=3in]{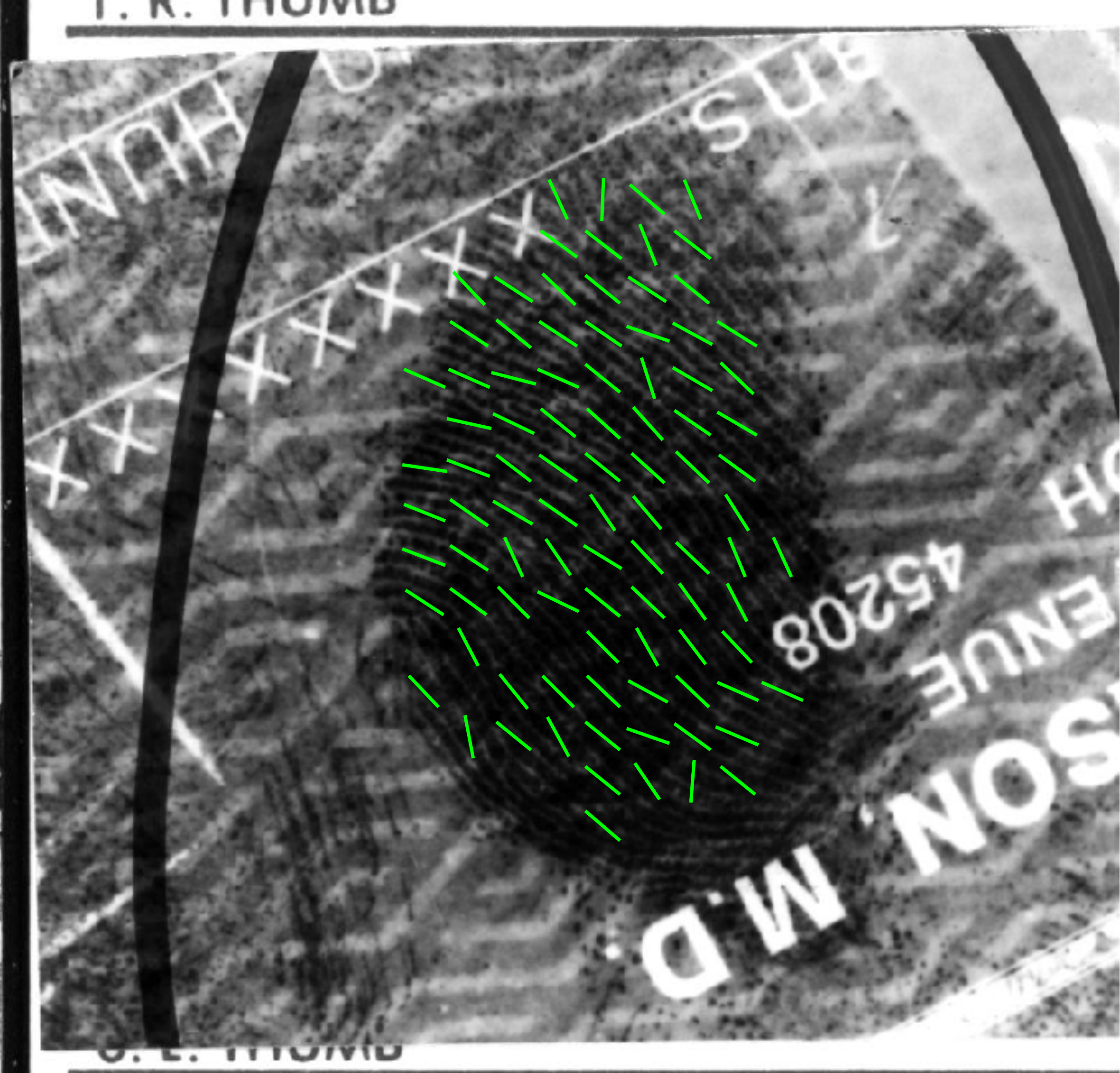}
        \label{latent_OF_modeling:latent_coarse_OF}
    }
    \subfigure[]
    {
        \includegraphics[width=3in]{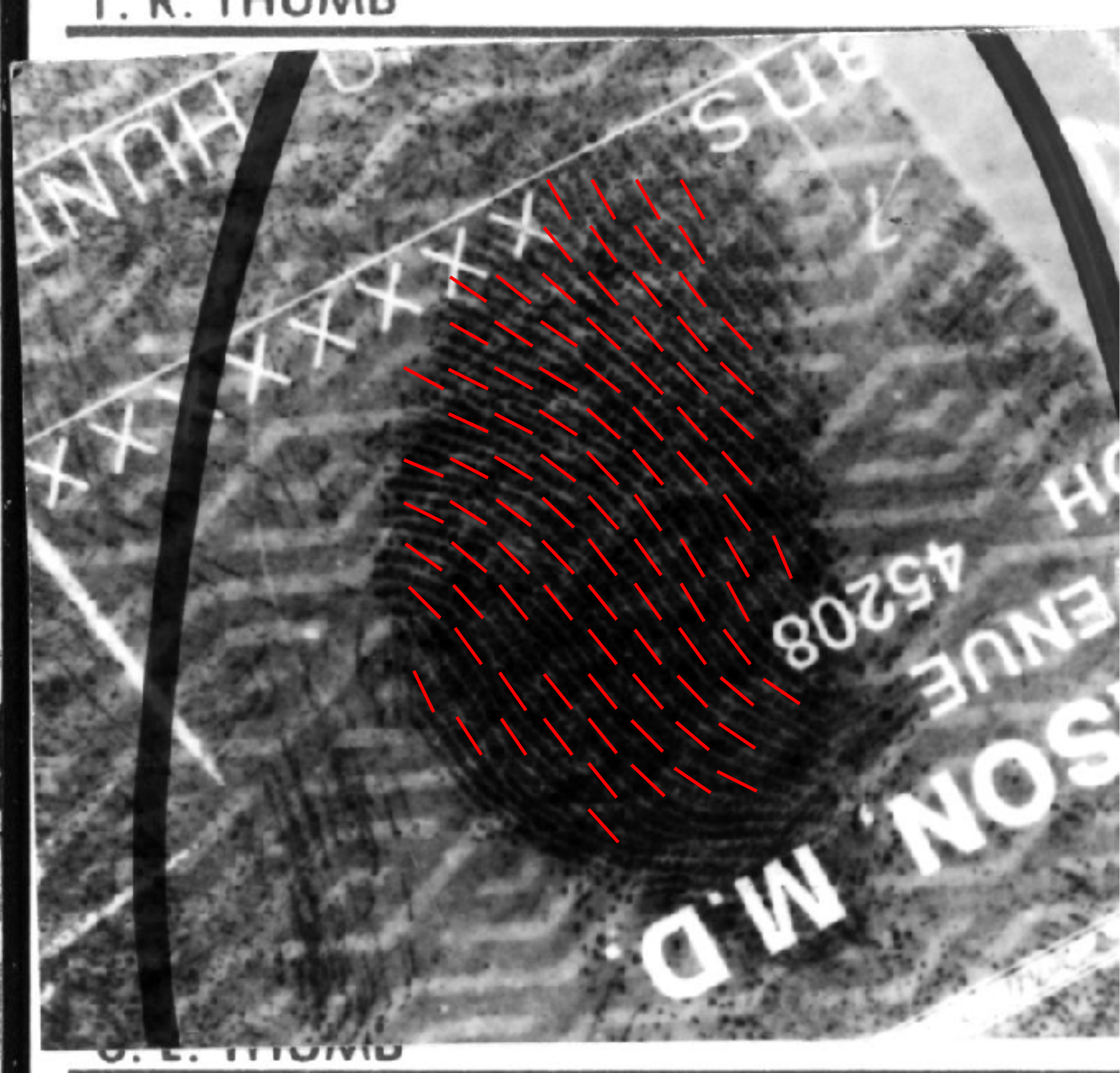}
        \label{latent_OF_modeling:latent_classical_FOMFE_OF}
    }
    \subfigure[]
    {
        \includegraphics[width=3in]{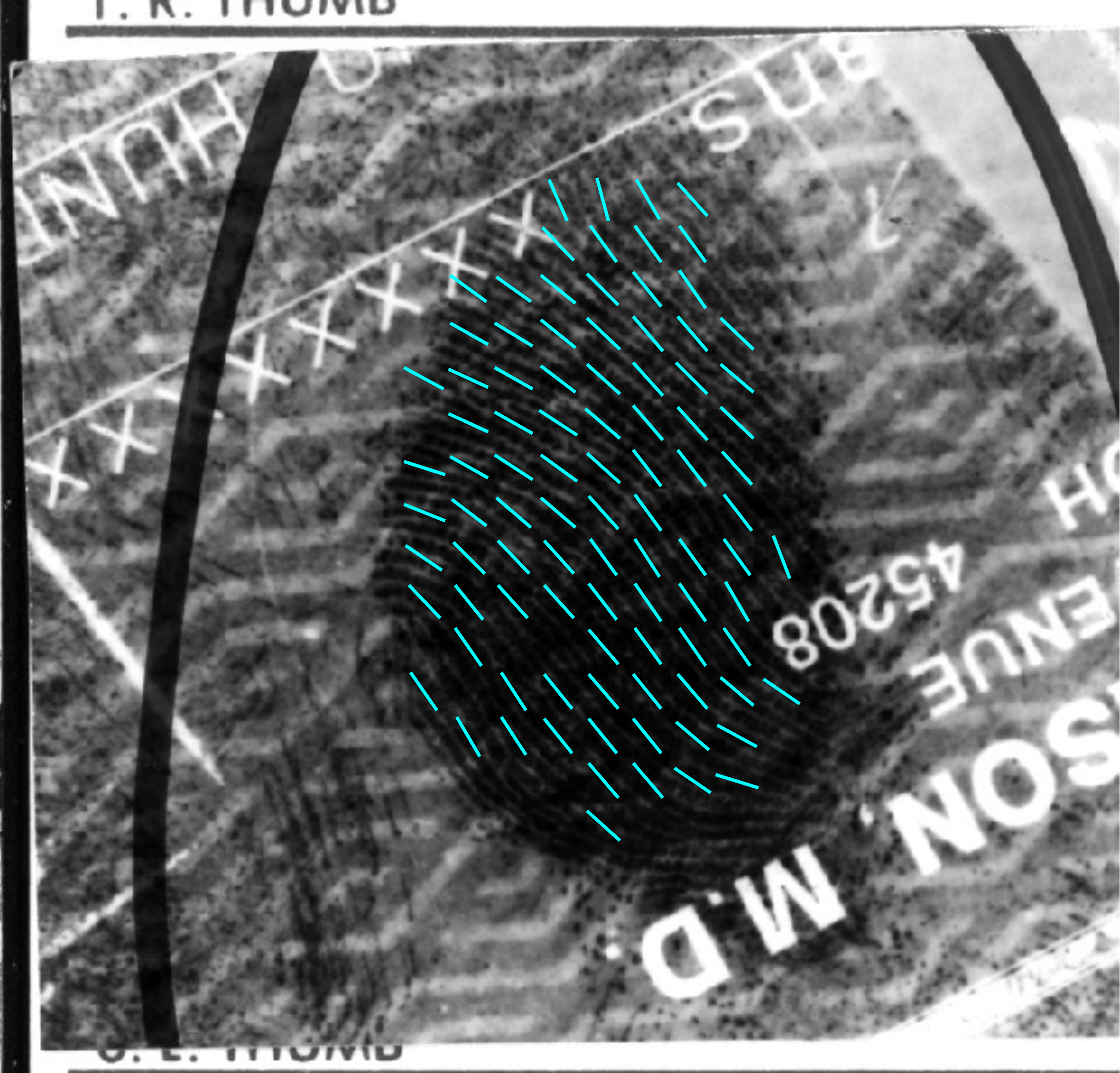}
        \label{latent_OF_modeling:latent_sparse_FOMFE_OF}
    }
    \subfigure[]
    {
        \includegraphics[width=3in]{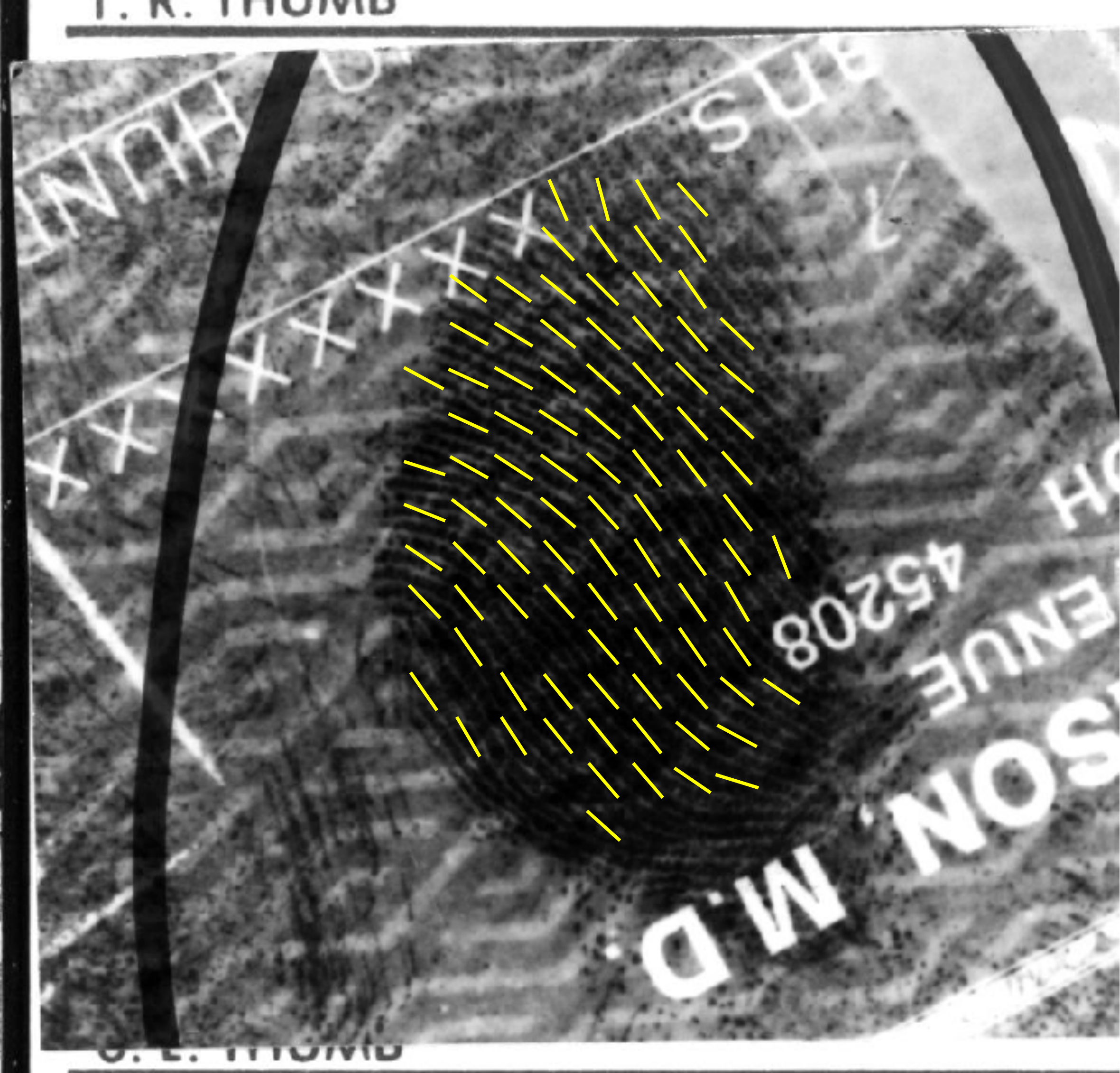}
        \label{latent_OF_modeling:latent_compressed_sparse_FOMFE_OF}
    }
    \caption{The illustration of coarse OF, classical FOMFE OF, s-FOMFE OF and cs-FOMFE OF for latent print image respectively: (a) coarse OF estimated by \cite{Maltoni2009}; (b) classical FOMFE OF; (c) s-FOMFE OF and (d) cs-FOMFE OF.}
    \label{latent_OF_modeling}
\end{figure}

As guided by the salient ridge map and its corresponding spatial mask, the coarse OF estimation approach is only applied to the image regions where the ridge structures are distinct and recognizable. Like the experiments on plain and rolled fingerprint, the same OF modeling parameters are employed to the latent print image. As a consequence, the constructed OFs via three different FOMFE models are illustrated in Figure \ref{latent_OF_modeling}. The results demonstrate that the proposed FOMFE models are also capable to model the OF for low-quality latent print images. For more reliable OF reconstruction, the image preprocessing in terms of salient ridge structure labeling is recommended to be performed prior to the OF modeling phase.

\section{Conclusion}

In this paper, two FOMFE-based OF modeling methods: s-FOMFE and cs-FOMFE are introduced by integrating with the rapidly developed signal sparse representation and compressed sensing theories. Such two models are able to model the OF for various types of fingerprints. The experiments on plain, rolled and latent fingerprint images demonstrate that the feasibility and effectiveness of the proposed models when modeling the ridge structure OF in a sparse or even a compressed sparse way. Further, the experimental results suggest that the proposed s-FOMFE and cs-FOMFE models can be adopted in the fingerprint indexing tasks under big-data circumstance. Future works will be conducted under the following two directions: (i) the application of s-FOMFE and cs-FOMFE models in latent fingerprint indexing tasks and (ii) the development of new sparse and compressed sparse OF models by exploiting diverse basis (like Legendre polynomials basis \cite{Ram2010} and DCT basis \cite{Liu2014}) instead of FOMFE basis.

\section*{Acknowledgment}

The first author Jinwei Xu would like to thank Wei Zhou at The University of New South Wales, Canberra Campus, for her helpful discussion.



\end{document}